\documentclass[journal]{IEEEtran}
\usepackage{cite}
\ifCLASSINFOpdf
   \usepackage[pdftex]{graphicx}

\else

   \usepackage[dvips]{graphicx}

\fi

\usepackage{amsmath}

\usepackage{algorithmic}

\usepackage{array}
\usepackage{fixltx2e}
\usepackage{stfloats}

\ifCLASSOPTIONcaptionsoff
 \usepackage[nomarkers]{endfloat}
\let\MYoriglatexcaption\caption
\renewcommand{\caption}[2][\relax]{\MYoriglatexcaption[#2]{#2}}
\fi

\usepackage{xspace}
\usepackage{multirow}
\usepackage{hyperref}
\usepackage{subfigure}
\usepackage{color}
\usepackage{algorithm}
\usepackage{booktabs}

\newcommand{\adaptfu}{SimAdapter\xspace}
\newcommand{\metaad}{MetaAdapter\xspace}
\newcommand{\adaptall}{SimAdapter+\xspace}

\begin{document}

\title{Exploiting Adapters for Cross-lingual Low-resource Speech Recognition}

\author{Wenxin~Hou,
        Han~Zhu, % Equal Contribution with Yidong， alphabet order
        Yidong~Wang,
        Jindong~Wang,
        Tao~Qin,
        Renjun~Xu,
        and~Takahiro~Shinozaki% <-this % stops a space~\IEEEmembership{Life~Fellow,~IEEE}
\thanks{W. Hou, Y. Wang, and T. Shinozaki are with Tokyo Institute of Technology, Tokyo, Japan. Email: \{hou.w.aa, wang.y.ca\}@m.titech.ac.jp, shinot@ict.e.titech.ac.jp.}% <-this % stops a space
\thanks{W. Hou is also with Microsoft, work done at Tokyo Institute of Technology and Microsoft Research Asia. Email: wenxinhou@microsoft.com}
\thanks{H. Zhu is with Institute of Acoustics, Chinese Academy of Sciences, China. Email: zhuhan@hccl.ioa.ac.cn.}% <-this % stops a space
\thanks{J. Wang and T. Qin are with Microsoft Research Asia. Email: \{jindong.wang, taoqin\}@microsoft.com.}% <-this % stops a space
\thanks{R. Xu is with Zhejiang University. Email: rux@zju.edu.cn.}% <-this %
\thanks{Correspondence to Jindong Wang and Takahiro Shinozaki.}
}


\markboth{IEEE Transactions on Audio, Speech, and Language Processing}%
{Hou \MakeLowercase{\textit{et al.}}: Bare Demo of IEEEtran.cls for IEEE Journals}

\maketitle

\begin{abstract}
 Cross-lingual speech adaptation aims to solve the problem of leveraging multiple rich-resource languages to build models for a low-resource target language. Since the low-resource language has limited training data, speech recognition models can easily overfit. Adapter is a versatile module that can be plugged into Transformer for parameter-efficient learning. In this paper, we propose to use adapters for parameter-efficient cross-lingual speech adaptation. Based on our previous \metaad that implicitly leverages adapters, we propose a novel algorithm called \adaptfu for explicitly learning knowledge from adapters. Our algorithms can be easily integrated into the Transformer structure. \metaad leverages meta-learning to transfer the general knowledge from training data to the test language. \adaptfu aims to learn the similarities between the source and target languages during fine-tuning using the adapters. We conduct extensive experiments on five-low-resource languages in the Common Voice dataset. Results demonstrate that \metaad and \adaptfu can reduce WER by 2.98\% and 2.55\% with only 2.5\% and 15.5\% of trainable parameters compared to the strong full-model fine-tuning baseline. Moreover, we show that these two novel algorithms can be integrated for better performance with up to 3.55\% relative WER reduction.
\end{abstract}


% \fi
\IEEEpeerreviewmaketitle

\section{Introduction}
\IEEEPARstart{A}{utomatic} speech recognition (ASR) based on end-to-end (E2E) models has made remarkable progress by training on large-scale data~\cite{wang2019overview, karita2019comparative}.
We can use a single E2E ASR system for a large number of languages~\cite{pratap2020massively, Hou2020} without complicated language-specific processing.
Nevertheless, a well-known limitation of E2E ASR methods is that they require a considerable amount of training data to achieve superior performances among various domains~\cite{hou21b_interspeech} and languages~\cite{conneau2020unsupervised}.
Therefore, it remains a challenge for E2E ASR models to achieve good performance for most of the low-resource languages among about 7,000 languages in the world.

\begin{figure}[t!]
\centering
\includegraphics[width=.48\textwidth]{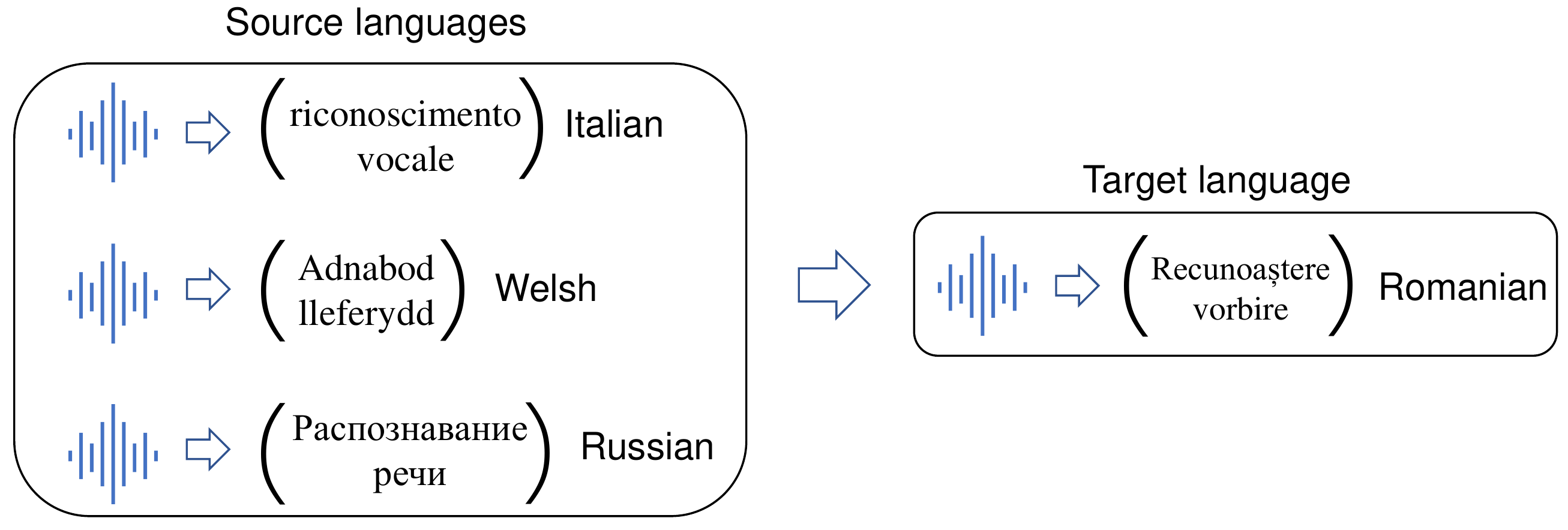}
\caption{Illustration of the cross-lingual speech recognition task. Given three rich-resource languages as source languages (Italian, Welsh, and Russian), how to learn the transferable knowledge from them to build cross-lingual ASR models for the target language Romanian?}
\label{fig-motiv}
\end{figure}

Some research has indicated that the performances of low-resource languages benefit from transferring the common knowledge from rich-resource languages in ASR~\cite{chuangsuwanich2016multilingual}. 
For instance, as shown in \figurename~\ref{fig-motiv}, given Romanian as a low-resource target language, cross-lingual ASR aims to build models by leveraging the available rich-resource languages such as Italian, Welsh, and Russian as source languages.
Knowledge transfer can be achieved in three avenues: (1) pre-training on the rich-resource languages and then fine-tuning on the low-resource tasks~\cite{pratap2020massively, adams2019massively}; (2) performing multi-task training using all languages~\cite{zhou2018multilingual}; and (3) learning the general common knowledge and then rapidly adapting to the low-resource languages using meta-learning~\cite{hsu2020meta}.
A possible explanation is that different languages intrinsically share some beneficial information like speaker, environment and linguistic information.
In this paper, we mainly focus on the fine-tuning methods.
%Kannan et al.~\cite{kannan2019large} proposed to oversample low-resource languages and use Adapter modules to effectively use the shared information.  

Due to the limited training data in low-resource languages, direct re-training makes the model easily overfit.
These problems make the transfer-based methods inefficient~\cite{hou2021meta,kannan2019large}.
Recently, the adapter module was proposed for parameter-efficient fine-tuning in multilingual or cross-lingual settings~\cite{kannan2019large, winata2020adapt, hou2021meta}, which can mitigate overfitting.
Adapter is an add-on module to the encoder and decoder layers in Transformer mainly composed of layer normalization and fully-connected layers.
During fine-tuning, we can freeze the backbones of the pre-trained models and only train the adapters which have a small number of task-specific parameters.
Pfeiffer et al.~\cite{pfeiffer2020adapterfusion} studied the fusion of adapters in natural language processing where they linearly combine the outputs of multiple adapters for target classification task adaptation.
However, it remains unexplored to investigate the performance of multiple adapters on cross-lingual ASR tasks.

In our previous work~\cite{hou2021meta}, we proposed \emph{\metaad} to learn general and transferable speech representations using model-agnostic meta-learning (MAML)~\cite{finn2017model} and achieved promising results on extremely low-resource languages.
However, it is unclear whether \metaad can handle the non-extreme cases where there are moderate training data. Moreover, \metaad relies on meta-learning to implicitly learn from source languages, which makes no assumptions on the relationship between source and target languages that may weaken its interpretability.
Therefore, in this paper, we comprehensively investigate the potential of leveraging multiple source adapters in cross-lingual speech recognition. Based on our analysis on \metaad, we propose a novel algorithm: \emph{\adaptfu}, to learn the similarity between the source and target languages using the attention mechanism. Our key motivation is that different languages in the world are sharing similarities based on their similar geological characteristics or evolution~\cite{ke2006language,macneilage2010origin,frayer200014}. 
Therefore, it is feasible to explicitly model such similarities in the ASR models.

Both of the two algorithms are parameter-efficient and thus can prevent the overfitting problem.
To our best knowledge, there is no existing research that tries to model the cross-lingual ASR tasks by studying their relationship using meta-learning and transfer learning-based adapters.
In addition, the \metaad and \adaptfu are compatible, thus can be integrated for better performance.

Our contributions can be summarized as follows:
\begin{itemize}
    \item We comprehensively analyze our previously proposed \metaad and propose a novel algorithm for cross-lingual low-resource ASR: \adaptfu.
    \item Experiments on five low-resource languages demonstrated a relative WER improvement of 2.98\% with \metaad and 2.55\% with \adaptfu using only 2.5\% and 15.5\% trainable parameters compared with the strong full-model fine-tuning baseline, respectively.
    \item These two algorithms can be integrated to achieve better performance with up to 3.55\% relative improvement.
\end{itemize}

This paper is substantially an extended version of our previously published paper~\cite{hou2021meta} at ICASSP 2021. Compared to the previous version, we make heavy extensions as follows: (1) We propose a parallel new algorithm called \adaptfu for cross-lingual ASR. (2) We investigate the difference and integration between the \metaad and \adaptfu algorithms. (3) We conduct extensive experiments on cross-lingual ASR datasets to validate the effectiveness of these algorithms.

The structure of this paper is as follows.
In Section~\ref{sec-related}, we review the related work to multilingual, cross-lingual ASR and adapters.
Section~\ref{sec-overview} introduces our main ideas.
Section~\ref{sec-meta} and Section~\ref{sec-fusion} introduce the details of \metaad and \adaptfu algorithms, and their integration.
Section~\ref{sec-exp} presents experimental design details and Section~\ref{sec-expresult} reports our experimental results and analysis.
Finally, in Section~\ref{sec-con}, we conclude this paper and present some future work.

\section{Related Works}
\label{sec-related}

\subsection{Multilingual and Cross-lingual Speech Recognition}
% Multilingual ASR
Multilingual E2E ASR is getting increasing attention over the years to handle multiple languages with a single model.
Watanabe et al.~\cite{watanabe2017language} proposed a language-independent architecture based on hybrid CTC-attention structure~\cite{kim2017joint} with augmented vocabulary for character-based E2E ASR and joint language identification. Toshniwal et al.~\cite{toshniwal2018multilingual} found that multilingual training leads to a significant relative improvement of recognition performance and the results can be further boosted by conditioning the model on a language identifier. Some attempts take a step towards realizing language-universal ASR.
Li et al.~\cite{li2019bytes} proposed to replace the characters with the Unicode bytes as the output.
Datta et al.~\cite{datta2020language} unified different writing systems through a many-to-one transliteration transducer. Recently, large-scale multilingual ASR systems have been investigated~\cite{kannan2019large, adams2019massively, Hou2020, pratap2020massively, li2021scaling}. \cite{pratap2020massively} proposed jointly training on 16,000 hours of speech data of 51 languages with up to 1 billion parameters. Inspired by~\cite{watanabe2017language}, Hou et al. presented LID-42~\cite{Hou2020}, a large-scale multilingual acoustic Transformer model trained on 11 mixed corpora of 42 languages.

% Cross-lingual ASR
Cho et al.~\cite{cho2018multilingual} validated the effectiveness of cross-lingual transfer learning for improving ASR performance. And this advantage can be further revealed by large-scale pre-training~\cite{adams2019massively, pratap2020massively}. For example, LID-42 can achieve a relative WER reduction of up to 28.1\% on low-resource languages by cross-lingual transfer~\cite{Hou2020}. Yi et al.~\cite{yi2018adversarial} introduced an adversarial learning objective to learn language-agnostic features. They appended a language discriminator after the shared encoder to distinguish which language the bottleneck features belong to. The objective of the discriminator is to correctly identify the language while the adversarial objective of the encoder is to fool the discriminator. The adversarial training process is realized with the use of the gradient reversal layer (GRL)~\cite{ganin2016domain}. 
Adams et al.~\cite{adams2019massively} performed experiments to analyze the impacts of language similarity, context-independent phoneme CTC objective and the aforementioned language-adversarial classification objective during multilingual pre-training to encourage language-agnostic features for better cross-lingual adaptation.

Other than learning the language-agnostic features, the optimization-based meta-learning approaches~\cite{finn2017model, nichol2018first} that aim to find a proper initialization for rapid adaptation have also been explored for cross-lingual ASR~\cite{hou2021meta}. Hsu et al.~\cite{hsu2020meta} proposed to apply the model-agnostic meta-learning (MAML)~\cite{finn2017model} as the pre-training method and achieved significant improvement over the conventional multilingual pre-training baseline. Xiao et al.~\cite{xiao2020adversarial} proposed the Adversarial Meta Sampling framework by introducing a policy network (task sampler) to dynamically sample languages based on their task difficulty. The ASR model is trained to minimize the loss while the task sampler learns to choose the most difficult languages that can maximize the loss. As a consequence, the learned initialization has a more balanced distance to all languages and shows a good generalization
capacity in low-resource speech tasks.

% Recently, some unsupervised pre-training methods like CPC and wav2vec 2.0 have also been explored in the multilingual setting~\cite{kawakami2020learning, conneau2020unsupervised}

\subsection{Adapters}
Due to the large quantity of parameters contained in the Transformer-based models~\cite{devlin2019bert, baevski2019vq, Hou2020, baevski2020wav2vec}, recent literature proposed the \emph{Adapter} structure~\cite{houlsby2019parameter, bapna2019simple} for parameter-efficient adaptation of pre-trained Transformers~\cite{vaswani2017attention, devlin2019bert} on various downstream tasks including language understanding~\cite{wang2018glue} and neural machine translation (NMT)~\cite{vaswani2017attention}, etc.
Adapter is a versatile module that can be plugged into the Transformer blocks.
The general philosophy for adapter-based fine-tuning is to freeze the parameters $\theta_b$ of the Transformer backbone and only tune the parameters $\theta_a$ of the adapter.
Compared to fine-tuning the whole Transformer model, fine-tuning the adapters is significantly efficient with acceptable performance loss~\cite{houlsby2019parameter}. Therefore, adapters have been adopted as a fine-tuning technique in few-shot domain adaptation for NMT~\cite{sharaf2020meta} and unsupervised cross-lingual transfer~\cite{artetxe2020cross} or domain adaptation~\cite{li2020unsupervised} of large-scale pre-trained language models like BERT~\cite{devlin2019bert} and XLM~\cite{lample2019cross}. Li et al.~\cite{ye2021zero} proposed a hypernetwork that could generate parameters of task-specific adapters from task descriptions to enable zero-shot learning~\cite{weller2020learning}. 
More recently, \cite{pfeiffer2020adapterfusion} introduced AdapterFusion to fuse adapters trained on different tasks to share the knowledge. The difference between our work and theirs is that we focus on the cross-lingual sequence-to-sequence ASR task while they experiment on text classification based on BERT~\cite{devlin2019bert}.

Some researchers have proposed to apply the Adapters to the E2E ASR tasks. In~\cite{kannan2019large}, Kannan et al. proposes to use the adapters to handle the
data imbalance problem for large-scale multilingual ASR. After obtaining the model trained on the union of data from all languages, they trained the language-dependent adapters on each of the languages again so that the multilingual backbone shares information across languages while the adapters could allow for per-language specialization. Winata et al.~\cite{winata2020adapt} extends this idea by further introducing a common adapter for all languages to learn language-agnostic information in the multilingual data. On the other hand, Hou et al.~\cite{hou2021meta} investigates the possibility of applying adapters to cross-lingual ASR under the assumption that a large-scale pre-trained multilingual model~\cite{Hou2020} should have contained adequate general acoustic and linguistic knowledge and could be adapted to any unseen target language with moderate feature adaptation. Furthermore, they proposed to pre-train the adapters with meta-learning to obtain the \metaad that provides a proper initialization for fast adaptation.

\adaptfu is similar to Mixture of Expert (MoE)~\cite{kim2021scalable}.
MoE is often used to scale up the model size while retaining the computing efficiency.
Therefore, there are often many experts and the expert outputs are often ``sparsely activated'' by using routing layers.
In practice, only top-$k$ experts are selected where $k=1$ or $2$. Also, MoE is trained on large-scale data along with the whole model where the experts acquire specific knowledge by themselves while each adapter component in our \adaptfu is ``taught'', and the \adaptfu is applied to parameter-efficient adaptation to low-resource languages.
However, we believe that some idea behind MoE is helpful to us. For example, as we observe that SimAdapter could distract its attention when the number of languages increases. We could also apply a similar routing mechanism to our SimAdapter. We will leave this for our future exploration.

\section{Exploiting Adapters for Cross-lingual ASR}
\label{sec-overview}
\subsection{Problem Definition}
The goal of cross-lingual speech recognition is to transfer the knowledge from the existing languages to the new language.
Formally speaking, given $N$ rich-resource languages $\{S_1, S_2, \cdots S_N\}$, cross-lingual ASR aims at adapting the pre-trained model to an unseen target low-resource language $L_{T}$.
Each language $S_i$ is composed of the speech-text pairs and we typically use $X$ and $y$ to denote them, respectively, i.e., $S_i = \{X_j, y_j\}^{N_i}_{j=1}$, where $N_i$ is the total number of training data.
Also note that the target language is low-resource compared to the training languages, i.e., $N_T \ll N_i, \forall 1 \le i \le N$.

\subsection{Overview}

In this paper, we comprehensively investigate the potential of adapters to achieve parameter-efficient cross-lingual speech recognition.
On the one hand, the parameters of the adapters are the only trainable parameters
%\wyd{is the parameters? How about the parameters of the adapters modules are trainable, while the rest parameters are frozen}
in the model with the rest parameters frozen, which remains parameter-efficient;
on the other hand, the adapters module can also help reduce overfitting on the low-resource cross-lingual data.

\begin{figure}[t!]
    \centering
    \includegraphics[width=.4\textwidth]{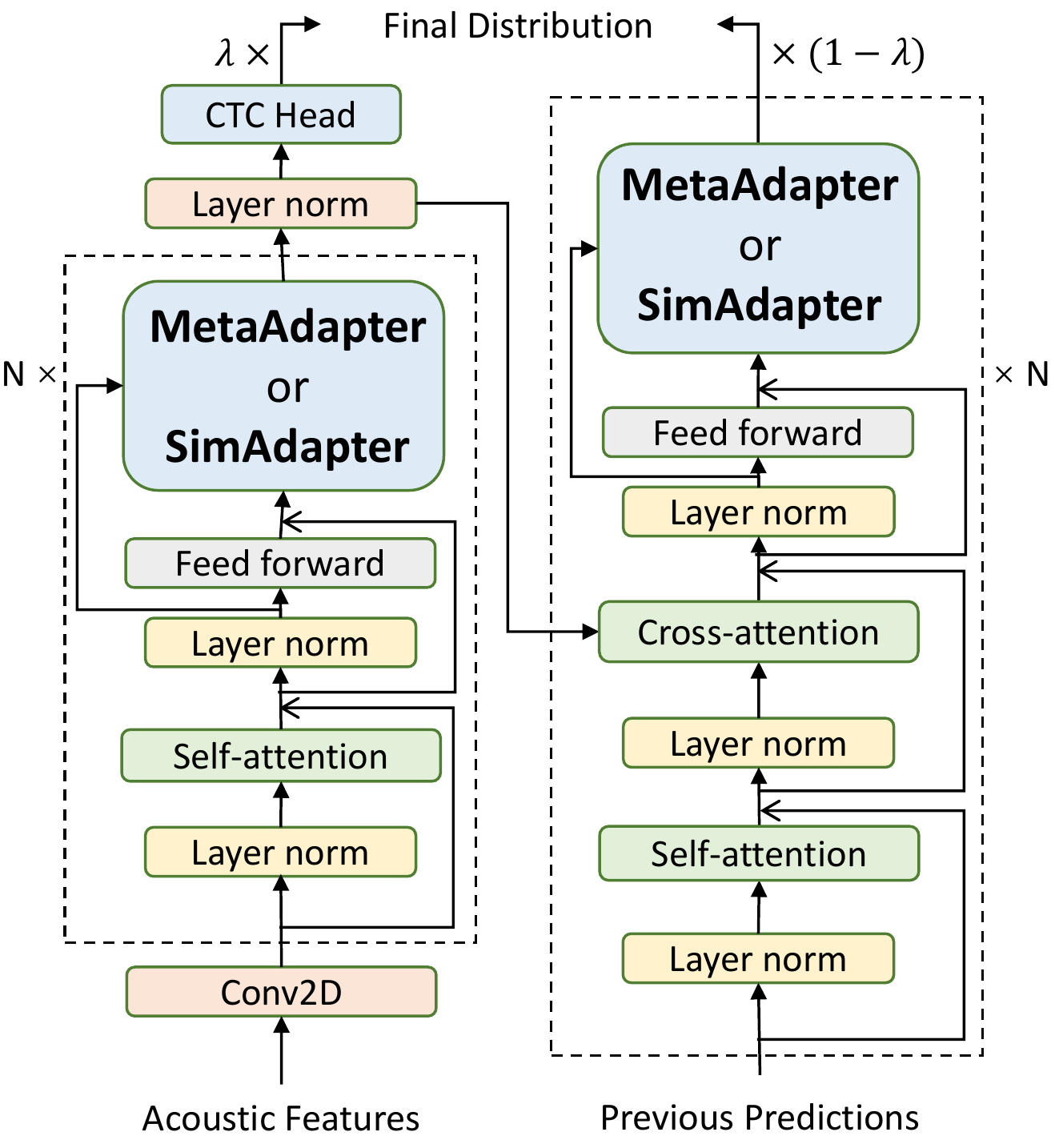}
    \caption{Illustration of the \metaad and \adaptfu module injected in the Speech-Transformer. Note that the residual connection between the feed-forward layer and layer normalization only applies to the \adaptfu.}
    \label{fig-transformer}
\end{figure}

To exploit adapters for cross-lingual ASR, it is important to study the relationship between different languages.
In this paper, we comprehensively analyze the \emph{\metaad} as well as the newly proposed \emph{\adaptfu} algorithms that learn and exploit the inter-language relationships to improve cross-lingual ASR.
Generally speaking, the \metaad is based on the meta-learning algorithm to extract general latent knowledge from existing training tasks and then adapt the knowledge to the target task. 
On the other hand, the \adaptfu algorithm is to directly explore the similarity between the source and target languages and then exploit such similarity to dynamically fuse the useful knowledge to the target language.
Finally, we show that these two algorithms are not independent, but can be integrated for better performance.
As shown in \figurename~\ref{fig-transformer}, our \metaad and \adaptfu can be easily plugged into the Transformer backbone for implementation.

\subsection{Backbone: Super Multilingual Transformer ASR Model}
\label{ssec-asr}
The super language-independent 42-lingual ASR model (LID-42) is proposed by Hou et al. in~\cite{Hou2020}. LID-42 is based on the \textit{big} Speech-Transformer~\cite{dong2018speech} and joint CTC-attention structure~\cite{kim2017joint}. We elaborate the model details below.

As model inputs, LID-42 accepts the 83-dimensional acoustic features (filter banks with pitch) computed with 10 ms frame shift and 25 ms frame length. The acoustic features are firstly subsampled by a factor of 4 by 2 convolution layers with kernel size 3 and stride 2. The resulted features have a receptive field of 100 milliseconds for each frame. Then the following encoder layers process the subsampled features by self-attention and feed-forward as illustrated in~\cite{vaswani2017attention}. Apart from self-attention and feed-forward, the decoder layers further accept the encoder outputs and perform cross-attention.

For the CTC-attention hybrid structure, an auxiliary CTC task~\cite{graves2006connectionist} is introduced for encoder outputs in order to encourage the monotonic alignment and accelerate the convergence speed~\cite{kim2017joint}. In training, a weighted sum of the sequence-to-sequence attention loss $\mathcal{L}_{\mathrm{ATT}}$ and the CTC loss $\mathcal{L}_{\mathrm{CTC}}$ is employed:
\begin{equation}
\label{eq:asrloss}
    \mathcal{L}_{\mathrm{ASR}} = (1 - \lambda) \mathcal{L}_{\mathrm{ATT}} + \lambda \mathcal{L}_{\mathrm{CTC}},
\end{equation}
where $\lambda$ denotes the weight of the CTC module. 

Similarly, during decoding, the CTC module outputs are used to re-score the beam search results of the Transformer decoder on-the-fly:
\begin{equation}
    \hat{Y} = \arg \max_{Y\in \mathcal{Y}} (1 - \lambda) \log P_{\mathrm{ATT}} (Y|X) + \lambda \log P_{\mathrm{CTC}}(Y|X),
\end{equation}
where $X$ are the 83-dimensional acoustic features (filter banks with pitch), $\mathcal{Y}$ denotes the set of the decoding hypotheses.

As model outputs, a shared vocabulary including characters/subwords and language tokens (e.g., {\ttfamily<en>}, {\ttfamily<fr>}) of 42 languages is adopted to realize language-independent training and recognizing. Furthermore, a language token is inserted to the beginning of each training label as an auxiliary language identification target. The model is trained to firstly identify the language before recognizing the speech contents. It is worth noting that we focus on monolingual transfer in this work. Therefore, language-specific heads are used and the language identification objective is dropped during fine-tuning.

LID-42 is trained on around 5000-hour labeled speech data mixing 11 corpora covering 42 languages and has revealed a strong performance on cross-lingual transfer learning tasks as shown in previous works~\cite{Hou2020, hou2021meta}.

\begin{figure}[htbp]
    \centering
    \includegraphics[width=.18\textwidth]{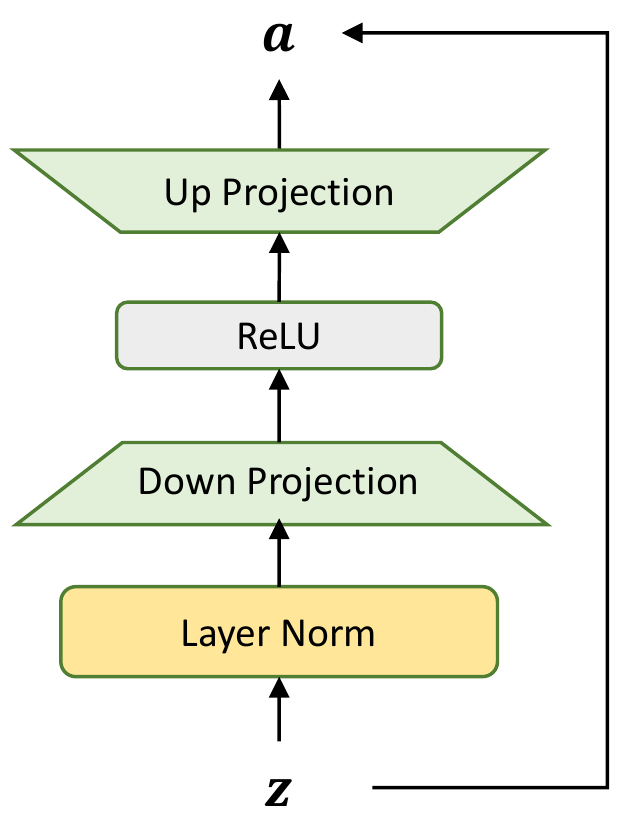}
    \caption{Architecture of the adapter module.}
    \label{fig-adapter}
\end{figure}

\subsection{Adapters}

As shown in \figurename~\ref{fig-adapter}, a commonly-used adapter structure includes layer normalization, a down-projection layer, a non-linear activation function, and an up-projection layer. There is also a residual connection that allows the adapter to keep the original representation unchanged. Thus, the adapter function is formulated as:
\begin{equation}
    \textbf{a}^l = \mathrm{Adapter}(\textbf{z}^l)= \textbf{z}^l + \textbf{W}_u^l \mathrm{ReLU}\left(\textbf{W}_d^l\left( \mathrm{LN}\left(\textbf{z}^l\right)\right)\right),
\end{equation}
where $\mathbf{z}^l$ represents the outputs of layer $l$, $\mathrm{LN}$ denotes layer normalization.
$\textbf{W}_u, \textbf{W}_d$ are weight parameters for up projection and down projection.

We will introduce these two algorithms and their integration in next sections.

\section{\metaad}
\label{sec-meta}
% \ts{This sub-section is not your proposed method, and is is for reviewing existing methods. Put this in separated section.}
% \hwx{There is minor modification in our method, i.e., we update only the adapter parameters in the model.}
% \ts{Separately explain it in your proposed method. Organization of text is important, where you should clearly separate exisiting methods and your proposed method.}
In this section, we introduce \metaad in detail.
\metaad is inspired by the idea of meta-learning~\cite{vanschoren2018meta} for fast adaptation to the new target tasks.
In our previous work~\cite{hou2021meta}, we investigated two meta-learning algorithms: Model-Agnostic Meta-Learning (MAML)~\cite{finn2017model} and Reptile~\cite{nichol2018first}. We found that MAML is more robust to the overfitting problem brought by the variance of adaptation data size and pre-training epochs. Therefore, we adopt the MAML as our meta-training algorithm in this work.

However, it is expensive to perform meta-learning directly on the full Speech-Transformer model since the model has heavy parameters that could easily overfit 
%on\wyd{remove on?}\hwx{Suggested by Grammarly? yes}
the low-resource target data.
To resolve this issue, \metaad utilizes the adapters to significantly reduce the adaptation parameters by learning aiming a proper initialization for faster adaptation.

\subsection{Architecture}

\begin{figure}[htbp]
    \centering
    \includegraphics[width=.4\textwidth]{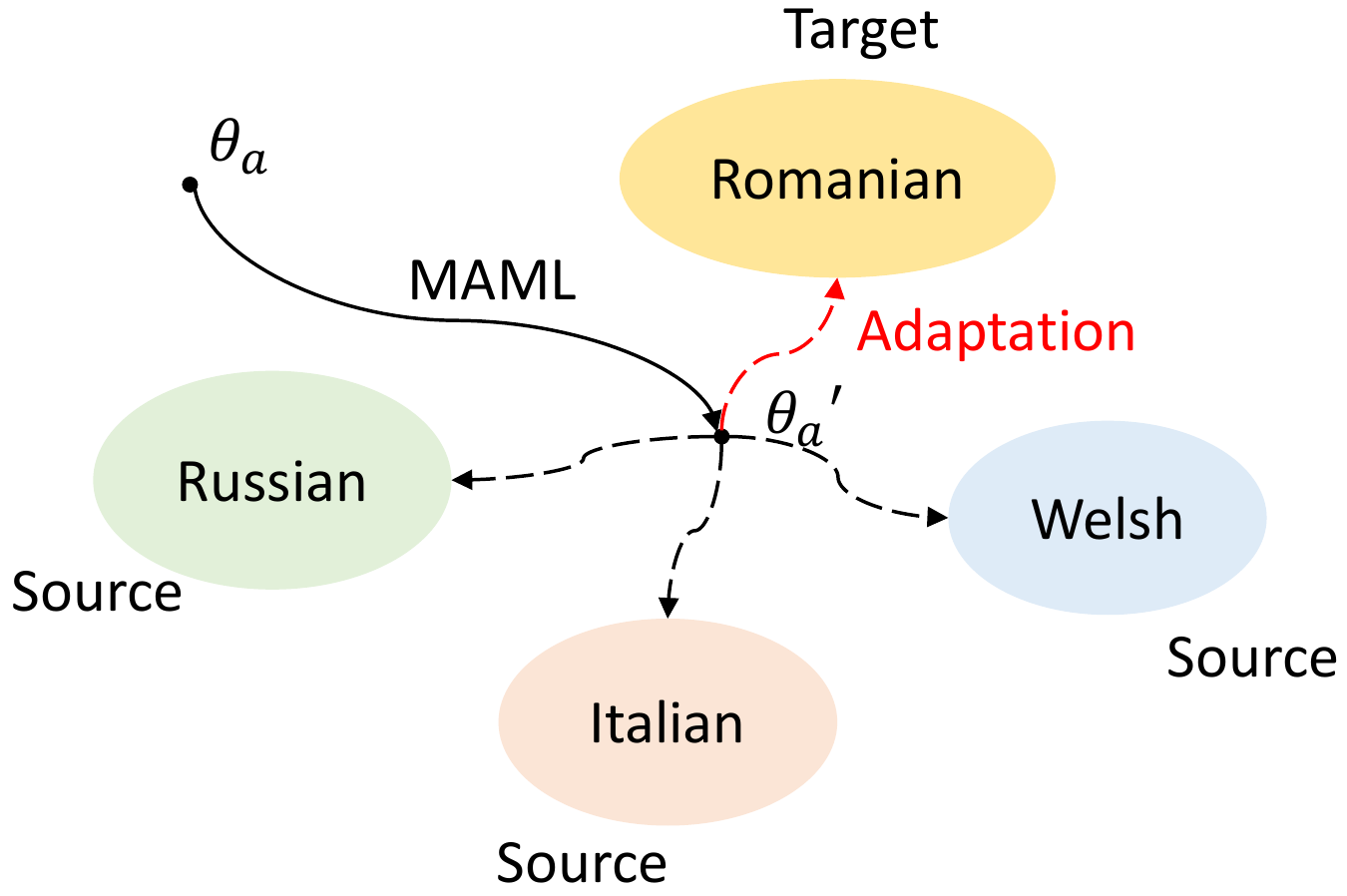}
    \caption{Illustration of \metaad.}
    \label{fig-metaad}
\end{figure}

The process of \metaad is illustrated in \figurename~\ref{fig-metaad}.
Given a pre-trained backbone speech-Transformer ASR model, \metaad is composed of two phases: (i) meta-train the \metaad on a bunch of source tasks; (ii) fine-tune the pre-trained adapter on unseen target tasks.

To use meta-learning, we view different languages as different tasks.
We split the parameters of \metaad into two types: the backbone parameters $\theta_b$ (i.e., vanilla Transformer) and the parameters of all adapters $\theta_a$.
Thus, given $N$ different source languages $\{S_1,S_2,\cdots,S_N\}$, we pre-train the \metaad module $f_{\theta_{a}}$ to obtain good initialization parameters $\theta_{a}$ that could generalize for fast adaptation given any unseen target language.
%The pre-trained meta-adapter is expected to adapt fast in the target language $T_i$ sampled from $m$ unseen target languages $\{T_1,T_2,T_3,\cdot\cdot\cdot,T_m\}$. 
Meanwhile, parameters of the pre-trained backbone $\theta_b$ are frozen during both the pre-training and the fine-tuning.

\subsection{Training \metaad}
In each pre-training episode, two subsets are randomly sampled from each source training language $S_i$, namely meta-training set $S^{tr}_i$ and meta-validation set $S^{val}_i$, i.e., $S^{tr}_i \cap S^{val}_i=\emptyset$. One episode is composed of two iterations: an inner iteration and an outer iteration. In the inner iteration, MAML updates the adapter parameters $\theta_{a}$ by performing one or more gradient descent on $S^{tr}_i$.
For notation simplicity, the updated parameter for language $S_i$ using the inner gradient descent iteration is:
\begin{equation}
    \theta^\prime_{a,i}=\theta_a - \epsilon \nabla \mathcal{L}_{S^{tr}_i}\left(f_{\theta_a}\right),
\end{equation}
where $\mathcal{L}$ is the ASR loss function as introduced in section~\ref{ssec-asr} and $\epsilon$ is the fast adaptation learning rate. In the outer iteration, the adapter parameters are then optimized to improve the performance of $f_{\theta^\prime_{a,i}}$ with respect to $\theta_a$ across all the meta-validation sets $S^{val}_i$. The meta-optimization objective of the outer iteration is:
\begin{equation}
    \mathcal{L}_{S^{val}_i}(f_{\theta^\prime_{a,i}})=\mathcal{L}_{S^{val}_i} \left(f_{\theta_a -\epsilon \nabla_{\theta_a} \mathcal{L}_{S^{tr}_i}\left(f_{\theta_a}\right)}\right).
\end{equation}

We optimize the meta-optimization objective through gradient descent as:
\begin{equation}
    \label{eq-maml}
    \theta_a =\theta_a - \mu  \sum_{i=1}^{N} \nabla_{\theta_a} \mathcal{L}_{S^{val}_i}\left(f_{\theta^\prime_{a,i}}\right),
\end{equation}
where $\mu$ is the meta step size.
 %There is a gradient through a gradient in the update procedure of the MAML.
 
After pre-training, the \metaad should obtain a proper initialization for any unseen target language(s). The complete training procedure of the \metaad is presented in Algo.~\ref{algo-meta}.

\begin{algorithm}[htbp]
	\caption{Learning algorithm of the \metaad}
	\label{algo-meta}
	\textbf{Input}: Rich-resource languages $\{S_1, \cdots, S_N\}$, low-resource task $L_{T}$.
	
% 	\textbf{Output}: $\{\Phi^\ast\}$.
	\begin{algorithmic}[1] %[1] enables line numbers
	    \STATE Train language-specific heads on source languages $S_i$.
		\STATE Initialize the \metaad.
		\WHILE{meta-learning not done}
		%\STATE 
		\STATE Optimizing the \metaad using Eq.~\eqref{eq-maml}.
		\ENDWHILE
		\STATE Train the target head on target language $L_T$.
		\STATE Fine-tune the \metaad using ASR loss Eq.~\eqref{eq:asrloss}.
		\STATE \textbf{return} Cross-lingual ASR model.
	\end{algorithmic}
\end{algorithm}

\section{\adaptfu}
\label{sec-fusion}
We propose \adaptfu to improve the adapter-based cross-lingual adaptation as well as the model interpretability by explicitly leveraging the knowledge of the source languages from the adapter modules. Here, `Sim' refers to similarity. 
% \subsection{Overview}
% \label{sec-method-overview}

% Common approaches include fine-tuning the whole pre-trained model on the target language. However, an obvious limitation is that full-model fine-tuning could easily get overfit due to the large volume of Transformer parameters versus the small-scale training data.
% In addition, it is also time-consuming and expensive to fine-tune and maintain the whole Transformer model parameters.
% Therefore, previous work~\cite{hou2021meta} has turned to using the adapter structure for parameter-efficient fine-tuning. 
% Nevertheless, there is still an obvious performance gap between the adapter modules and the conventional full-model fine-tuning. 

% \ts{Some of the description here is redundant with the introduction. Reducing the redundancy will help improve the section organization. This section should focus on concrete technical description. Overview thinks should be explained in the introduction.}
%We propose to use \adaptfu~\cite{pfeiffer2020\adaptfu} for cross-lingual ASR tasks shown in \figurename~\ref{fig-transformer}.

\adaptfu is inspired by existing research on language and speech origins~\cite{ke2006language,macneilage2010origin,frayer200014}, which implies that different languages in the world are sharing similarities based on their similar geological characteristics or cultural developments. 
%\ts{"are" is duplicated in "world are are sharing"}
Thus, it is feasible to leverage the knowledge of multilingual adapters for new target languages.

% Concretely speaking, \adaptfu leverages the attention mechanism to learn the similarities between source and target languages, which will be introduced in Section~\ref{sec-method-fusion}.
% Moreover, we propose a novel fusion guide loss to avoid the attention distraction problem to facilitate training in Section~\ref{sec-method-guide}.

%Adapters are separately trained and their knowledge are not be shared, which reduces the efficiency of the model with the increasing of target tasks or the number of adapters. To handle this problem, Pfeiffe et al.~\cite{pfeiffer2020\adaptfu} introduced the \adaptfu method and improved the adapter-based adaptation performance by fusing the knowledge from multiple tasks.\adaptfu module leverages the attention mechanism and allows the model to dynamically activate and fuse task-specific knowledge to generate better representations for adaptation. Motivated by the idea of adapters, \adaptfu is designed to be a non-destructive method and there is no catastrophic forgetting problem.%

\begin{figure}[t!]
    \centering
    \includegraphics[width=.4\textwidth]{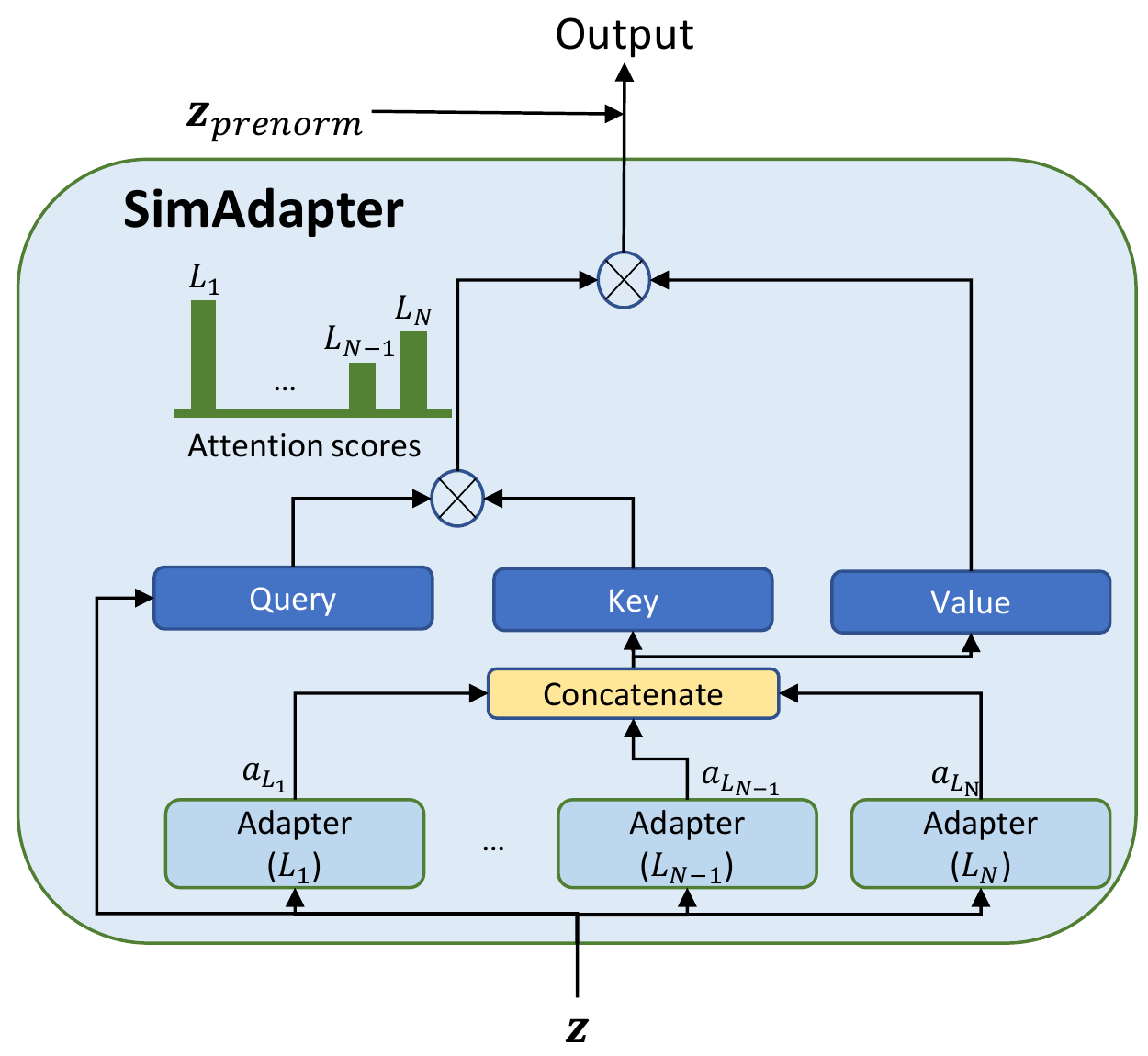}
    \caption{Illustration of the \adaptfu module. The language-specific features $a_{L_k}$ of different languages $L_k \in \{L_1, L_2, ..., L_N\}$ are attended by the language-agnostic features $\textbf{z}$ to extract better features for the target language.
%    \ts{Can you include explanation about $z$, $L_k$, and $AF$ in caption?} \hwx{Would this be okay now?} \ts{Yes}
    }
    \label{fig-fusion}
\end{figure}

\subsection{Architecture}
\label{sec-method-fusion}
\adaptfu is a parameter-efficient algorithm that learns the similarities between existing language-specific adapters and the target low-resource language based on the attention mechanism~\cite{vaswani2017attention}. 
Similar to the adapters, \adaptfu can also be easily integrated with existing pre-trained models and adapters.

% \adaptfu uses the attention mechanism~\cite{vaswani2017attention} to learn such similarities.
% Attention is widely used in sequence learning tasks by learning to attend to the most important areas or sequences.
% On the other hand, \adaptfu learns the attentions based on the adapters for parameter-efficient learning.

The detailed composition of the \adaptfu is shown in \figurename~\ref{fig-fusion}. By taking the language-agnostic representations from the backbone model as the query, and the language-specific outputs from multiple adapters as the keys and values, the final output for \adaptfu over attention are computed as (For notation simplicity, we omit the layer index $l$ below): 
% \begin{equation}
%     \textbf{A} = [\textbf{a}_{S_1}, \textbf{a}_{S_2}, ..., \textbf{a}_{S_N}]
% \end{equation}
\begin{equation}
    \mathrm{SimAdapter}(\mathbf{z}, \mathbf{a}_{\{S_1, S_2, ..., S_N\}})=\sum_{i=1}^N {\mathrm{Attn}(\mathbf{z}, \mathbf{a}_{S_i}})
    \cdot  \left( \mathbf{a}_{S_i} \mathbf{W}_V \right),
\end{equation}
where $\mathrm{SimAdapter}(\cdot)$ and $\mathrm{Attn}(\cdot)$ denotes the \adaptfu and attention operations, respectively.
Specifically, the attention operation is computed as:
\begin{equation}
    \mathrm{Attn}(\mathbf{z}, \mathbf{a})= \mathrm{Softmax} {\left(\frac{(\mathbf{z} \mathbf{W}_Q) (\mathbf{a}\mathbf{W}_K)^\top}{\tau}\right)},
\end{equation}
where $\tau$ is the temperature coefficient,  $\mathbf{W}_Q, \mathbf{W}_K, \mathbf{W}_V$ are attention matrices. Note that while $\mathbf{W}_Q, \mathbf{W}_K$ are initialized randomly, $\mathbf{W}_V$ is initialized with a diagonal of ones and the rest of the matrix with small weights ($1e-6$) to retain the adapter representations. Furthermore, a regularization term is introduced to avoid drastic feature changes:
\begin{equation}
    \mathcal{L}_\mathrm{reg} = \sum_{i, j} \left({(\mathbf{I}_V})_{i, j} - {(\mathbf{W}_V})_{i, j} \right)^2,
\end{equation}
where $\mathbf{I}_V$ is the identity matrix with the same size as $\mathbf{W}_V$. 
%\in (D, D)$. $z \in (T, D)$, $A \in (N, T, D)$, $Attn \in (T, N)$. $(T, D) (T, D) --> (T, N)$ 

In our cross-lingual setting, the \adaptfu module is expected to utilize language-specific knowledge from existing language adapters.

\subsection{Fusion Guide Loss}
\label{sec-method-guide}
%\hwx{Not good: For cross-lingual ASR tasks, the language-specific head layers need to be learned from scratch at the same time. However, our experiments show that the target adapter can get confused due to the strong language-specific heads focus too much on its own language that is hard to bridge the gap between source and target languages.}
Although \adaptfu aims to benefit from the similar knowledge of other languages, we believe that the most crucial information is stored in the adapter of the target language. However, since the weights of source and target adapters are initialized equally, \adaptfu often distracts its attention significantly from the target language during adaptation and generally does not perform well in our experiments. To alleviate this problem, we propose a fusion guide loss to encourage the model to focus on the corresponding adapters for the similarity learning. Specifically, for each language fusion layer $f$, we average the cross entropy of adapter attention scores among all $K$ time steps and ${S}$ samples.
The layer-wise guide losses are added up as:

\begin{equation}
    \mathcal{L}^f_\mathrm{guide} = -\frac{1}{K \times S}\sum_{s=1}^{S}\sum_{k=1}^{K}\log \alpha_{f, k}^{s},
\end{equation}
\begin{equation}
    \mathcal{L}_\mathrm{guide} = \sum_{f} \mathcal{L}^f_\mathrm{guide}.
\end{equation}
Note that $K$ represents the number of frames in the encoder and the number of tokens in the decoder side, $\alpha_{f, k}^{s}$ denotes the attention score of target language's Adapter. In this way, the attention scores are optimized via conventional Empirical Risk Minimization (ERM)~\cite{vapnik1998statistical}.

\subsection{Training \adaptfu}
\label{sec-method-train}
A difference between the previous application of AdapterFusion~\cite{pfeiffer2020adapterfusion} and our \adaptfu for cross-lingual ASR is that a language-specific language head is required to be trained for the unseen target language. However, training the Adapters together with the language heads may result in the insufficient learning of semantic information in the adapters. Therefore, in this work, we introduced a three-stage training strategy for \adaptfu-based ASR cross-lingual adaptation.

In the first stage, different from the previous work~\cite{hou2021meta}, \adaptfu trains the language-specific heads for each source language $S_i$ as well as the target language separately. This step aligns the language heads to the same latent semantic space of the backbone model. In the second stage, adapters are trained based on the pre-trained heads to learn the information. In the third stage, \adaptfu leverages the fusion of source languages for better adaptation to the target language. Only the parameters of the \adaptfu are trained in this stage.

By adding the $\mathbf{W}_V$ regularization term weighted by $\eta$ and the fusion guided loss weighted by $\gamma$, the final adaptation objective is given by:
\begin{equation}
    \label{eq-all}
    \mathcal{L} = \mathcal{L}_\mathrm{ASR} + \eta \mathcal{L}_\mathrm{reg} + \gamma \mathcal{L}_\mathrm{guide}.
\end{equation}

The complete training procedure of \adaptfu is presented in Algorithm~\ref{algo}.

\begin{algorithm}[htbp]
	\caption{Learning algorithm of \adaptfu}
	\label{algo}
	\textbf{Input}: Rich-resource languages $\{S_1, \cdots, S_N\}$, low-resource task $L_{T}$.
	
% 	\textbf{Output}: $\{\Phi^\ast\}$.
	\begin{algorithmic}[1] %[1] enables line numbers
	    \STATE Train language-specific heads on the source languages 
	    %$(X_{S_i}, Y_{S_i})$
	    $S_i$ and the target language.
		\STATE Train Adapters $A_t$ on top of language-specific heads.
		\STATE Initialize \adaptfu layers.
		\WHILE{not done}
		%\STATE 
		\STATE Optimizing \adaptfu layers using Eq.~\eqref{eq-all}.
		\ENDWHILE
		\STATE \textbf{return} Target ASR model.
	\end{algorithmic}
\end{algorithm}

\subsection{Integration of \metaad and \adaptfu}
Although \metaad and \adaptfu can both benefit cross-lingual adaptation by leveraging the knowledge of source languages, they are designed from different perspectives. \metaad aims to obtain a proper initialization for fast adaptation by learning from the source languages, which can be regarded as a type of latent transfer. On the other hand, \adaptfu explicitly models the similarities between source and target languages with the attention mechanism. Therefore, \metaad is good at handling extremely low-resource languages, while with more training data \adaptfu can capture the language similarities more precisely.

Moreover, note that \metaad and \adaptfu are not independent, but can be integrated into one algorithm, which we denote as \emph{\adaptall}.
We can simply fuse the source adapters with the target adapter learned by the \metaad using \adaptfu.
This can be seen as a two-stage knowledge transfer process where we aim to learn general and transferable knowledge from meta-learning in the first stage; then, we perform adaptation using the \adaptfu algorithm for fine-grained knowledge transfer to achieve better performance.

\section{Experimental Setup}
\label{sec-exp}
\subsection{Data Set}
We adopt the Common Voice 5.1~\cite{ardila2020common} corpus for our experiments. We follow the official data splits for training, validation and testing. For the \adaptfu, we select five rich-resource languages as source languages and five low-resource languages as targets.
Note that the source and target languages are all from European and some of them are spoken in geographically close regions to empirically analyze if the language similarities can be revealed by \adaptfu.
The detailed data statistics are shown in \tablename~\ref{tab-data}.

\begin{table}[htbp]
\caption{Training / validation / testing hours}
\label{tab-data}
\centering
\begin{tabular}{crrrr}
\toprule
& Language     & Train & Valid & Test  \\ \hline
\multirow{5}{*}{Source} & Russian (ru)           & 80.61 & 11.78 & 12.61 \\ 
& Welsh (cy)           & 74.84 & 4.35  & 4.25  \\ 
& Italian (it)           & 88.74 & 19.74 & 20.85 \\ 
& Basque (eu)           & 73.26 & 7.46  & 7.85  \\ 
& Portuguese (pt)           & 37.40 & 5.40  & 5.85  \\ \hline
\multirow{5}{*}{Target} & Romanian (ro)  & 3.04  & 0.42  & 1.66  \\ 
& Czech (cs)     & 20.66 & 2.84  & 3.13  \\ 
& Breton (br)    & 2.84  & 1.51  & 1.75  \\ 
& Arabic (ar)    & 7.87  & 2.01  & 2.09  \\ 
& Ukrainian (uk) & 17.35 & 2.30  & 2.36  \\ \bottomrule
\end{tabular}
\end{table}

\subsection{Compared Approaches}
We consider the following fine-tuning-based approaches as well as both end-to-end and conventional hybrid models and trained from random initialization for comparison. To evaluate the efficiency of different methods, we also list numbers of trainable parameters in Table~\ref{tab-param}.
It is shown that our \metaad and \adaptfu (and \adaptall) only use $\mathbf{2.5}\%$ and $\mathbf{15}\%$ of the training parameters from the full model, respectively, which are significantly parameter-efficient.
\begin{table}[htbp]
\centering
\caption{Comparison of number of trainable parameters.}
\label{tab-param}
\begin{tabular}{lr}
\toprule
Method & \# Trainable Parameters \\
\hline
Hybrid DNN/HMM & 14,247K \\ % 14246972
\hline
Full Model          & 27,235K \\ % 27,235,016 \\
Head          & 77K \\ % 77,000    \\
Head+(Meta-)Adapter       & 676K \\ % 676,040   \\
Head+(Meta-)Adapter+\adaptfu & 4,224K \\% 4,224,200 \\
\bottomrule
\end{tabular}
\end{table}

\begin{enumerate}
    \item Baselines without applying transfer learning:

\begin{itemize}
    \item DNN/HMM: Standard hybrid DNN/HMM models are trained with lattice-free MMI~\cite{povey2016purely} criterion using Kaldi~\cite{povey2011kaldi}. Specifically, we use 9 layers TDNN~\cite{peddinti2015time} the acoustic model. The acoustic features are 100-dimensional i-vector~\cite{dehak2010front} and 40-dimensional MFCC. 3-gram language model is applied for decoding.~\footnote{We did not find proper pronunciation dictionary for Breton. Therefore, only results of the other 4 languages are presented.}
    \item Trans.(B): We train a randomly-initialized big Transformer model of the same size and architecture as LID-42 from scratch.
    \item Trans.(S): To mitigate overfitting, we decrease the feed forward network from 2048 to 1024 so that the number of model parameters is reduced from 27,235K to 18,664K.
    \item Head: We keep the backbone model (LID-42) frozen as feature extractor and train the language-specific heads on top of it.
\end{itemize}
\item Fine-tuning based transfer:
\begin{itemize}
    \item Full-FT: We fine-tune the full model on each target language individually, leading to 5 separate models.
    \item Full-FT+L2: We apply L2 regularizations to Full-FT to avoid overfitting.
    \item Part-FT: We make only the last 3 decoder layers trainable and freeze the rest parameters to fine-tune on the target languages to mitigate overfitting. % There are around 8,856K trainable parameters.
\end{itemize}
\item Adapter based transfer:
\begin{itemize}
    \item Adapter: We inject and train the vanilla adapters while keeping the backbone model frozen.
    %\item Adapters (Trained with proposed strategy): we train the Adapters using our proposed training strategy.
    \item \metaad: We inject the pre-trained \metaad and train it as the vanilla adapters do.
    \item \adaptfu: We fuse the adapters of the source languages with the target language to improve the performance.
    \item \adaptall: Specifically, we combine the \metaad and the \adaptfu (namely \adaptall) to evaluate its performance and verify whether~\metaad and~\adaptfu are compatible.
\end{itemize}
\end{enumerate}

\subsection{Implementation Details}
We implement the E2E methods based on the ESPnet~\cite{watanabe2018espnet} codebase. The subword-based LID-42 model proposed in~\cite{Hou2020} is used as the backbone model for adaptation. The acoustic features are extracted following ESPnet. Numbers of SentencePiece~\cite{kudo2018sentencepiece} subwords are set to 150 and 100 for high- and low-resource languages, respectively.

We use Adam~\cite{kingma2014adam} as the optimizer.
For the full-model fine-tuning, we follow the same learning rate scheduling strategy as~\cite{vaswani2017attention} and warmup the learning rate to 0.2 in the first 10 epochs. The total number of training epochs is 200 for full-model fine-tuning and \adaptfu, and 100 for the other methods. 
Early stopping with patience 10 is adopted except for the training of source heads and adapters. 
The source languages heads and adapters are trained using a batch size of 1024 and a learning rate of 0.028. 
The target heads and adapters are trained using a batch size of 512 and a learning rate of 0.02.
For the \adaptfu, we use a batch size of 128 and a learning rate of $2e-5$.
%and \wyd{the} regularization loss weight 0.01\wyd{what does it mean? weight?}\hwx{yes, weight of reg loss}. 
%\hwx{Removed due to repetition}
We adopt $\eta=0.01$ for the regularization loss weight and 1.0 as the guide loss weight $\gamma$. The temperature coefficient $\tau$ is simply set to $1.0$.
We train the \metaad for 30 epochs using Adam~\cite{kingma2014adam} with $\beta_1=0$ in the inner training loop and vanilla stochastic gradient descent (SGD) in the outer loop. The inner adaptation learning rate and initial meta step size $\mu$ are 0.028 and 1.0, respectively. The meta step size linearly annealed to 0 over the course of training.
The weight of the CTC module $\lambda$ is set to 0.3 throughout the experiments following ESPnet~\cite{watanabe2018espnet}. Beam size 10 is employed for joint decoding.
Our source code is available at: \url{https://github.com/jindongwang/transferlearning/tree/master/code/ASR/Adapter}.

\subsection{Evaluation Metrics}
In this work, we use word error rate (WER) as our evaluation metric. We average the results on 5 languages to evaluate the overall performance of different methods by default. To reflect the performance on target languages according to their imbalanced test data duration (more test data often represents more training data), we also compute the weighted average WERs, which is more friendly to the methods that require relatively more training data. 
\begin{table*}[!t]
\centering
\caption{Word error rates (WER) on the cross-lingual ASR tasks}
\label{tab-main}
\resizebox{\textwidth}{!}{
\begin{tabular}{ccccc|ccc|cccccc}
\toprule
Target & DNN/HMM  & Trans.(B) & Trans.(S) & Head  & Full-FT & Full-FT+L2 & Part-FT & Adapter & \adaptfu & \metaad & \adaptall \\
\hline
Romanian (ro)     & 70.14 & 97.25  & 94.72    & 63.98 & 53.90 & 52.74   & 52.92 & 48.34   & 47.37              & \textbf{44.59}        & 47.29       \\
Czech (cs)        & 63.15 & 48.87 &  51.68     & 75.12 & 34.75 & 35.80  & 54.66 & 37.93  &  35.86              & 37.13        & \textbf{34.72}       \\
Breton (br)       & -     & 97.88  &  92.05   & 82.80  & 61.71 & 61.75  & 66.24 & 58.77   & \textbf{58.19}              & 58.47        & 59.14       \\
Arabic (ar)       & 69.31 & 75.32 &   74.88   & 81.70  & 47.63 & 50.09  & 58.49 & 47.31   &  47.23              & 46.82        & \textbf{46.39}       \\
Ukrainian (uk)    & 77.76 & 64.09  &  67.89   & 82.71 & \textbf{45.62}  & 46.45 & 66.12 & 50.84  &   48.73              & 49.36        & 47.41       \\
\hline
AVG          &   -    & 76.68   &  76.24  & 77.26 & 48.72  & 49.37 & 59.69 & 48.64   &  47.48              & 47.27        & \textbf{46.99}       \\
Weighted AVG &  -     & 72.28   &  72.50  & 77.54 & 46.72 & 47.50  & 59.43 & 47.38  & 46.08              & 46.12        & \textbf{45.45}      \\
\bottomrule
\end{tabular}}
\end{table*}

\section{Experimental Results}
\label{sec-expresult}

\subsection{Cross-lingual speech recognition}
Table~\ref{tab-main} shows the main results on cross-lingual ASR. 
The first three columns show the non-fine-tuning-based baselines. First, it can be found that the hybrid DNN/HMM model outperforms Transformer (big) on 2 out of 4 languages (Romanian (ro), Arabic (ar)), and these 2 languages are with least training data. The results indicate that the overfitting issue occurs in the Transformer model. Transformer (S) mitigates the problem to some extent but it is still far from satisfaction. It could further be inferred that even hybrid DNN/HMM has the overfitting problem on the extremely low-resource Romanian language, since lower WER is obtained with the linear head simply trained on top of the frozen but powerful LID-42 backbone.

On the other hand, from the fine-tuning- and adapter- based approaches presented on the middle- and right-hand sides, we can observe that the adapters successfully avoid the overfitting problem and outperform the Full-FT method on 3 very low-resource languages (Romanian, Breton, Arabic). Applying L2 regularization and partial fine-tuning both improve the performance on Romanian but degrades on the other 4 languages.
It can be also found that the~\metaad and~\adaptfu approaches can achieve similar and competitive results on the 5 target languages. 
Furthermore, we notice that both the~\metaad and~\adaptfu consistently improve the performance over the adapters and narrow the gap with Full-FT on the languages with relatively more training data (Czech and Ukrainian). Meanwhile, the~\metaad method performs better on the extremely low-resource languages (ar, ro) and has lower average WER, while~\adaptfu shows better results on moderate low-resource languages (br, cs) and obtains lower weighted average WER. 
Finally, by combining the~\metaad with~\adaptfu, the~\adaptall method surpasses all the other approaches and obtains the best average performance on the 5 languages, indicating that the two methods are compatible since they leverage the source information in different ways.
Combining the results from \tablename~\ref{tab-param} where \adaptall only uses 15.5\% trainable parameters of the full model, we see that \adaptall is both effective and parameter-efficient.

% \begin{table}[htbp]
% \centering
% \caption{Word error rates on the cross-lingual ASR tasks}
% \label{tab-main}
% \begin{tabular}{cccccc}
% \toprule
% Target      & Head   & Adapter & Full-FT & AF \\
% \hline
% ro  & 63.98  & 48.34   & 53.90    & \textbf{47.37}              \\
% cs     & 75.12  & 37.93   & \textbf{34.75}   & 35.86              \\
% br    & 82.80   & 58.77   & 61.71   & \textbf{58.19}              \\
% ar    & 81.70   & 47.31   & 47.63   & \textbf{47.23}              \\
% uk & 82.71  & 50.84   & \textbf{45.62}   & 48.73              \\
% \hline
% Average      & 77.26 & 48.64  & 48.72  & \textbf{47.48}        \\
% \bottomrule
% \end{tabular}
% \end{table}

\subsection{Ablation Study}
\subsubsection{Impact of different training strategies}
We compare the impact brought by different adapter-training strategies, i.e., jointly training the adapter with head and the first two stages of the training strategy proposed in Section~\ref{sec-method-train}. 
The results are presented in Table~\ref{tab-strategy}. 
It is clear that the proposed two-stage training strategy can consistently reduce the WERs of both the adapters and the \adaptfu.

\begin{table}[htbp]
\centering
\caption{Comparison of different Adapter training strategies.}
\label{tab-strategy}
\begin{tabular}{ccccc}
\toprule
Target             & Joint  & +\adaptfu  & Two-stage & +\adaptfu  \\
\hline
ro  & 52.92  & 53.88  & 48.34   & 47.37  \\
cs & 39.16  & 36.79  & 37.93   & 35.86  \\
br & 65.10   & 63.37  & 58.77   & 58.19  \\
ar & 50.53  & 49.31  & 47.31   & 47.23  \\
uk & 52.27  & 48.84  & 50.84   & 48.73  \\
\hline
Average & 52.00 & 50.44 & 48.64  & 47.48 \\
+Weighted & 50.35 & 48.57 & 47.38 & 46.08 \\
\bottomrule
\end{tabular}
\end{table}

\begin{figure}[htbp]
    \centering
    \includegraphics[width=.45\textwidth]{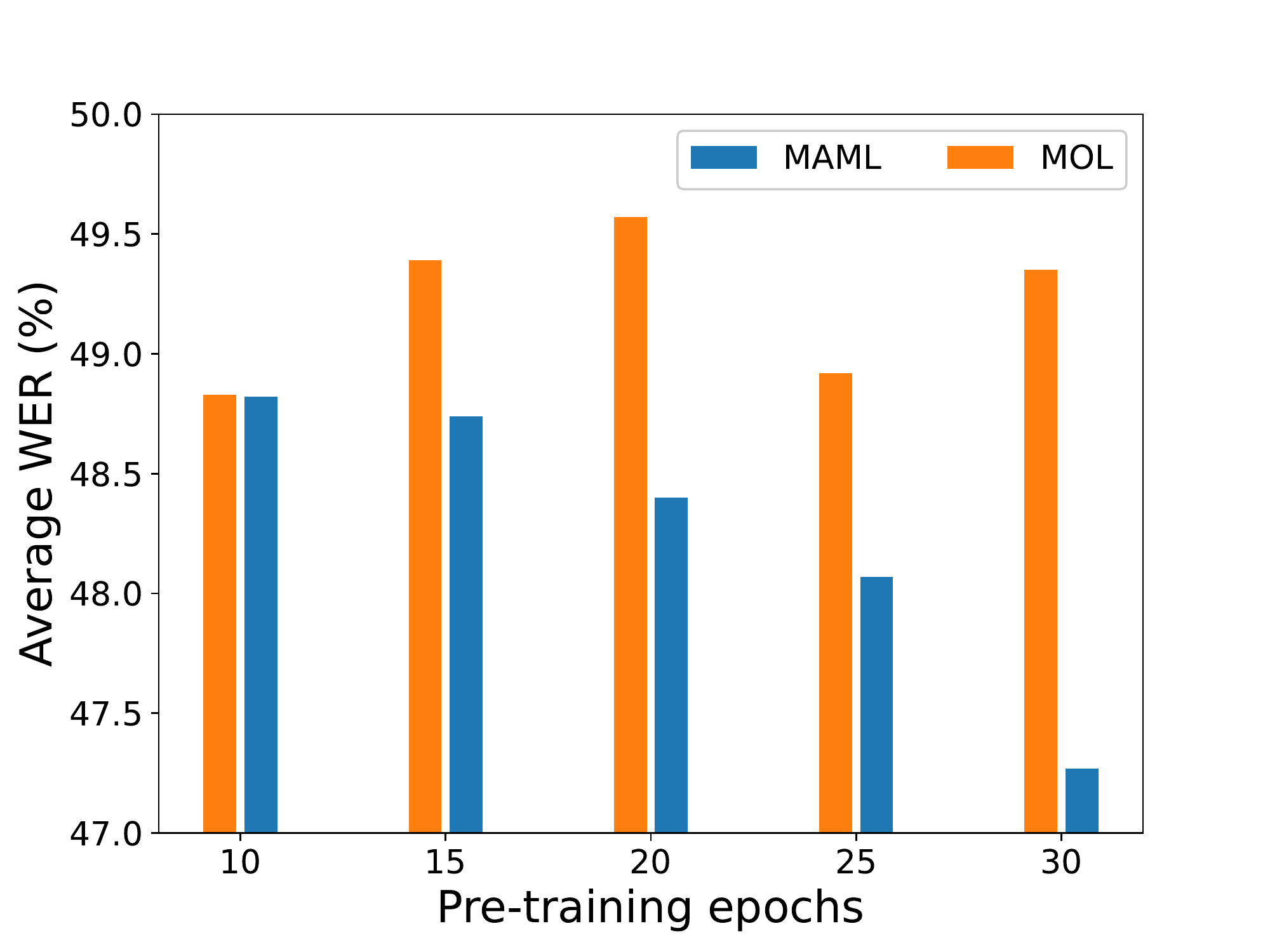}
    \caption{Comparison between MAML and conventional multi-objective learning (MOL) approach for Adapter pre-training.}
    \label{fig-pretrain-approach}
\end{figure}

\begin{figure*}[t!]
    \begin{minipage}[t]{1\linewidth}
        \centering
        \vspace{-.2in}
        %uk
        \subfigure[w/o target adapter (uk)]{\includegraphics[width=.24\textwidth]{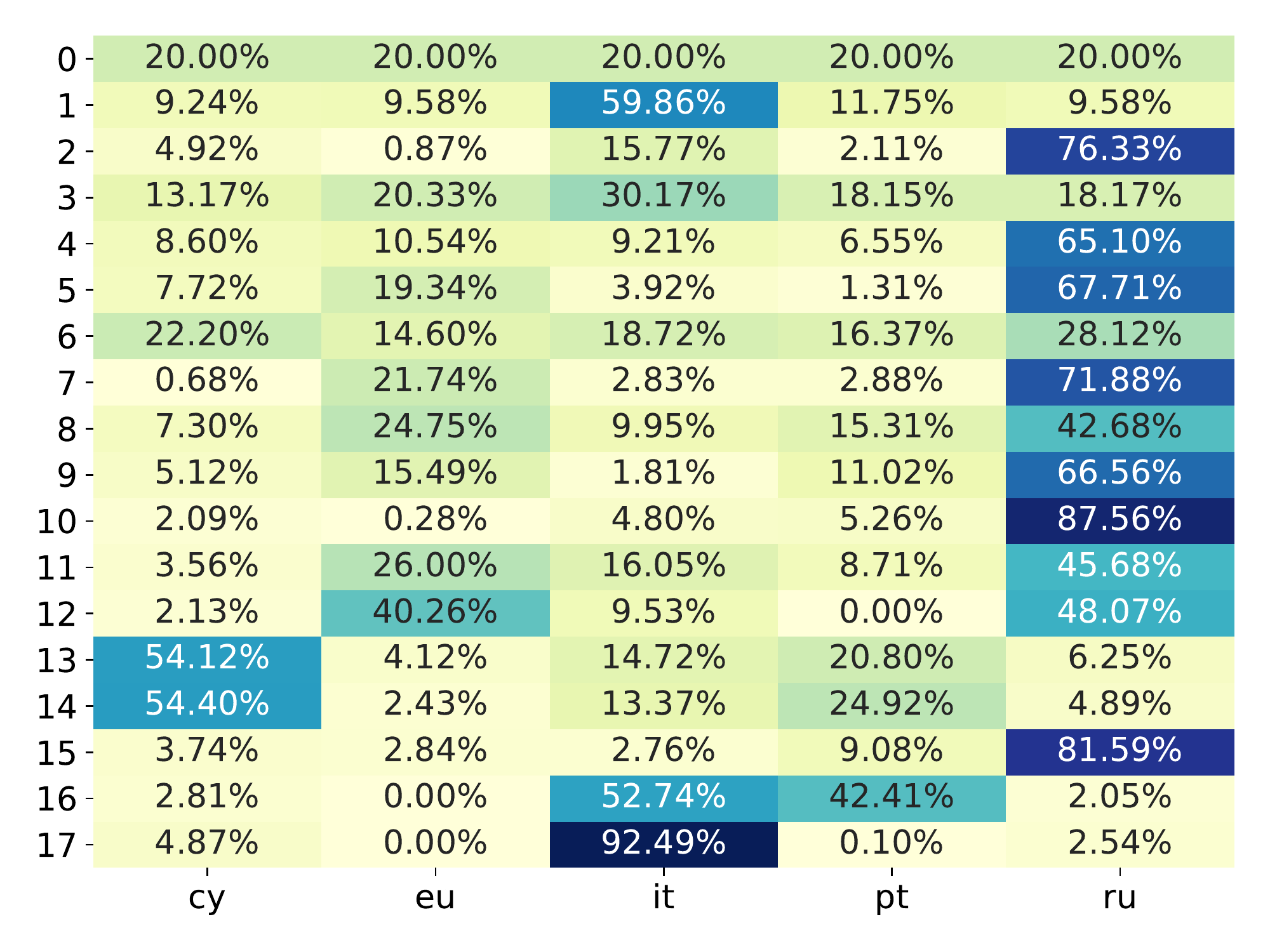}}
        \subfigure[w/o guide loss (uk)]{\includegraphics[width=.24\textwidth]{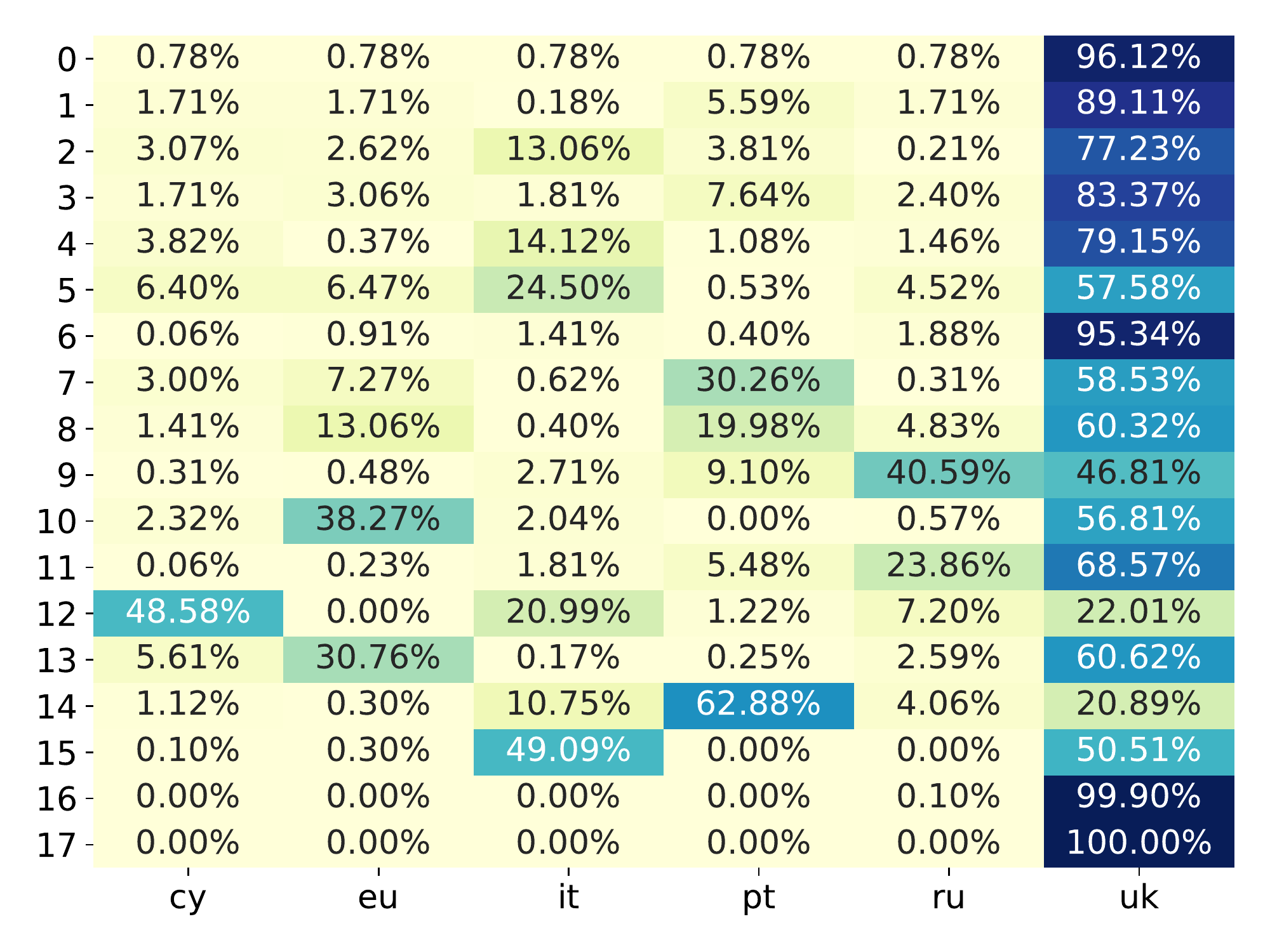}}
        \subfigure[target adapter + guide loss (uk)]{\includegraphics[width=.24\textwidth]{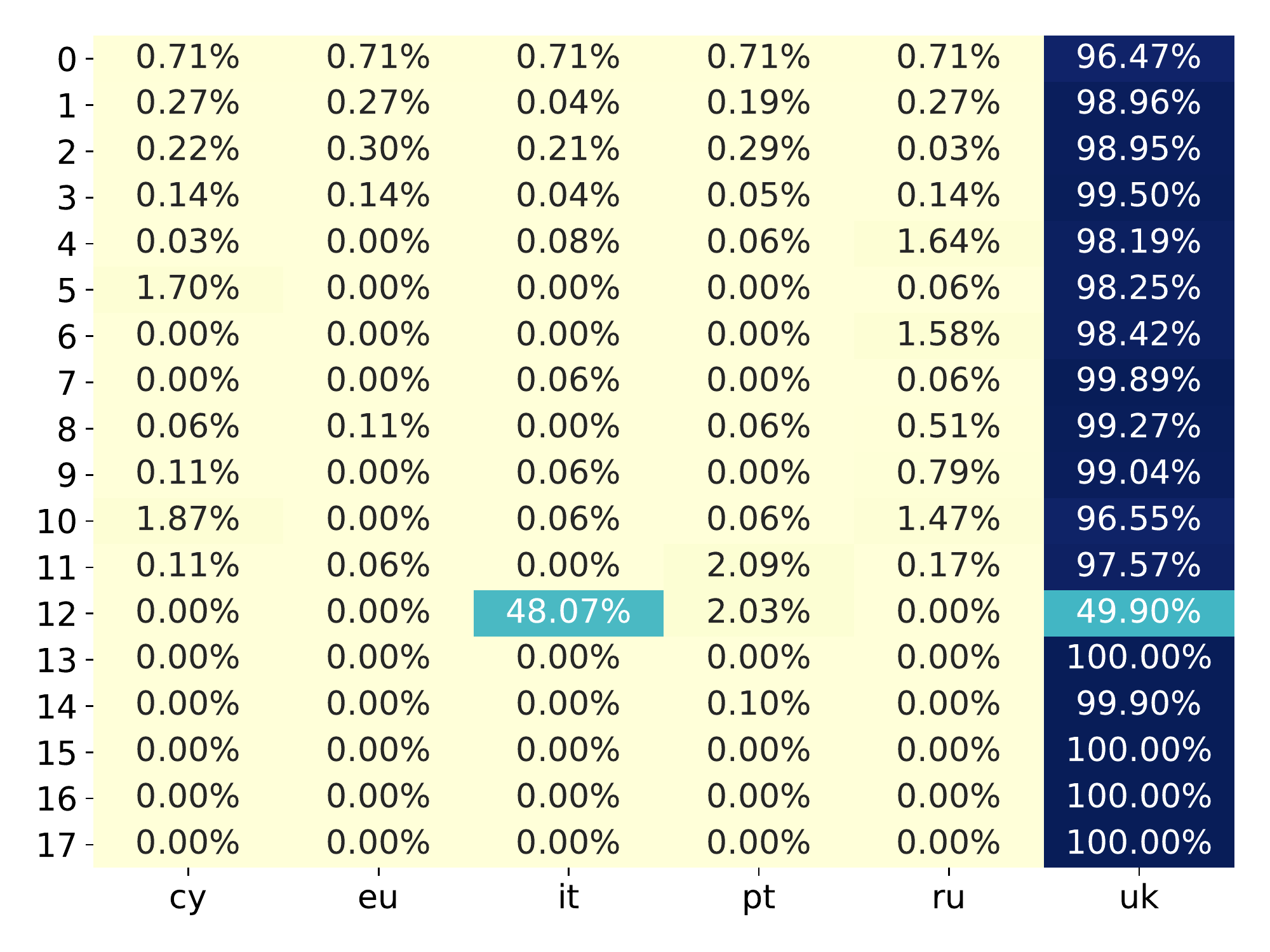}}
        \subfigure[meta-adapter + guide loss (uk)]{\includegraphics[width=.24\textwidth]{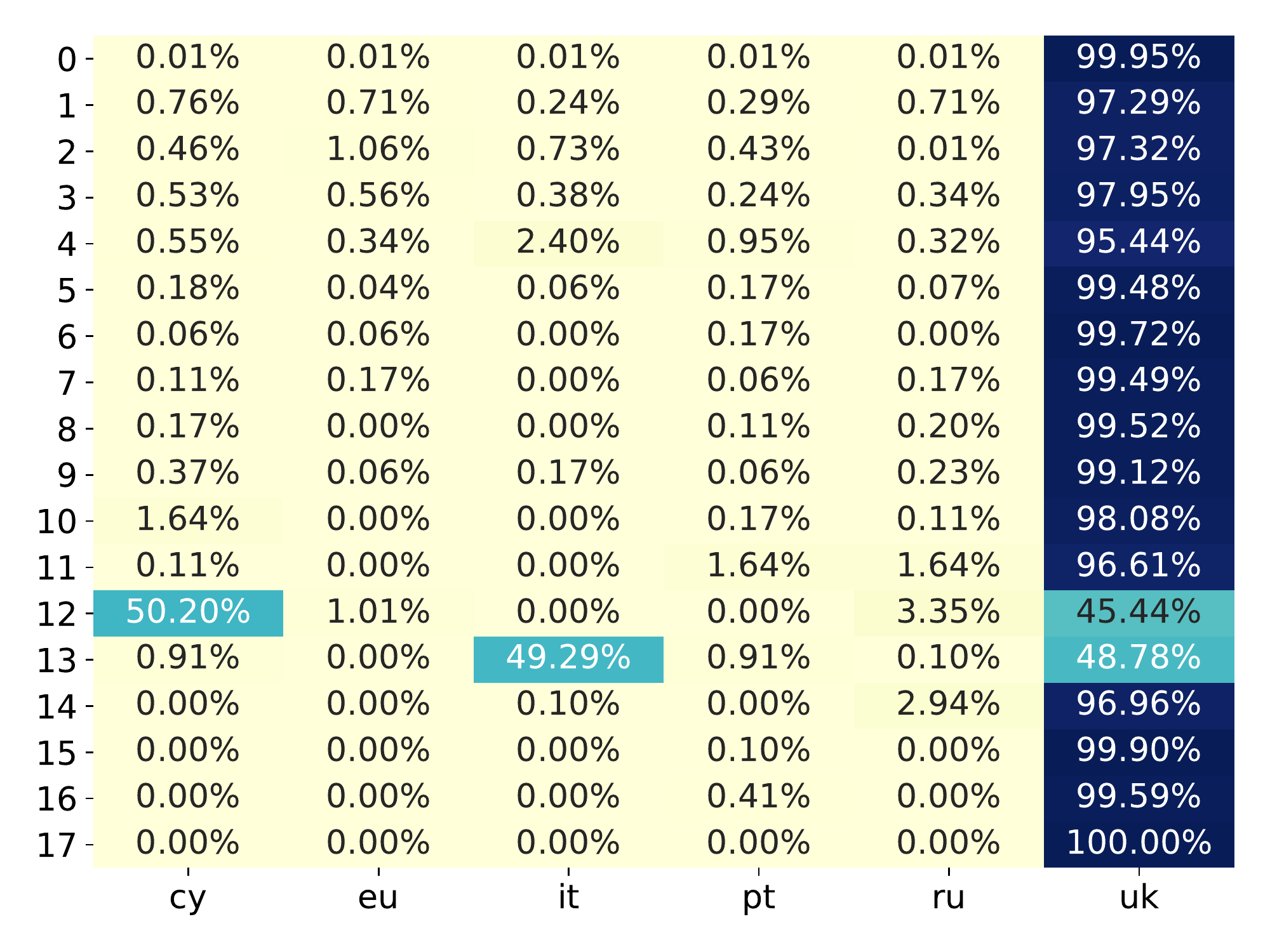}}
        %ar
        \subfigure[w/o target adapter (ar)]{\includegraphics[width=.24\textwidth]{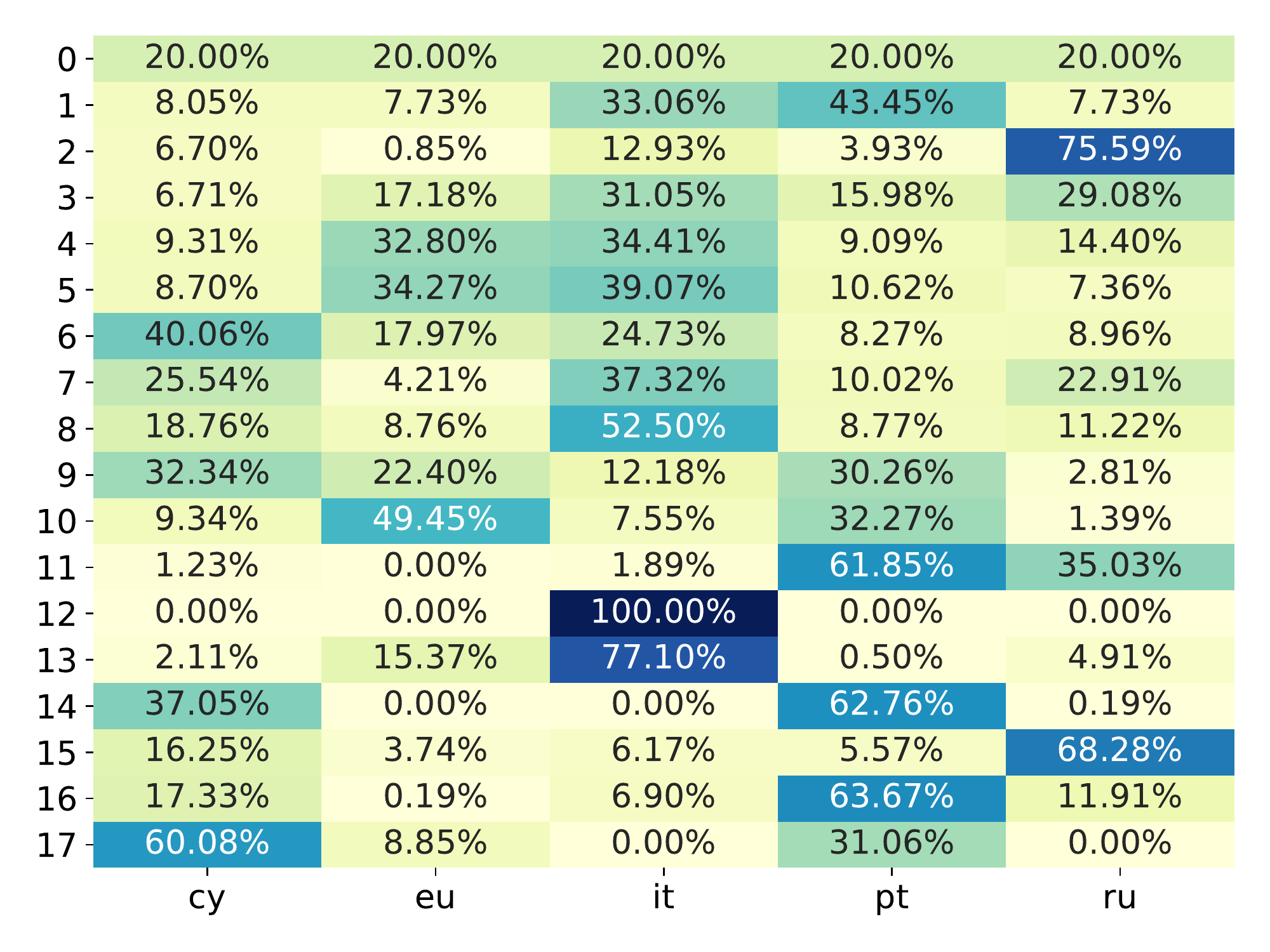}}
        \subfigure[w/o guide loss (ar)]{\includegraphics[width=.24\textwidth]{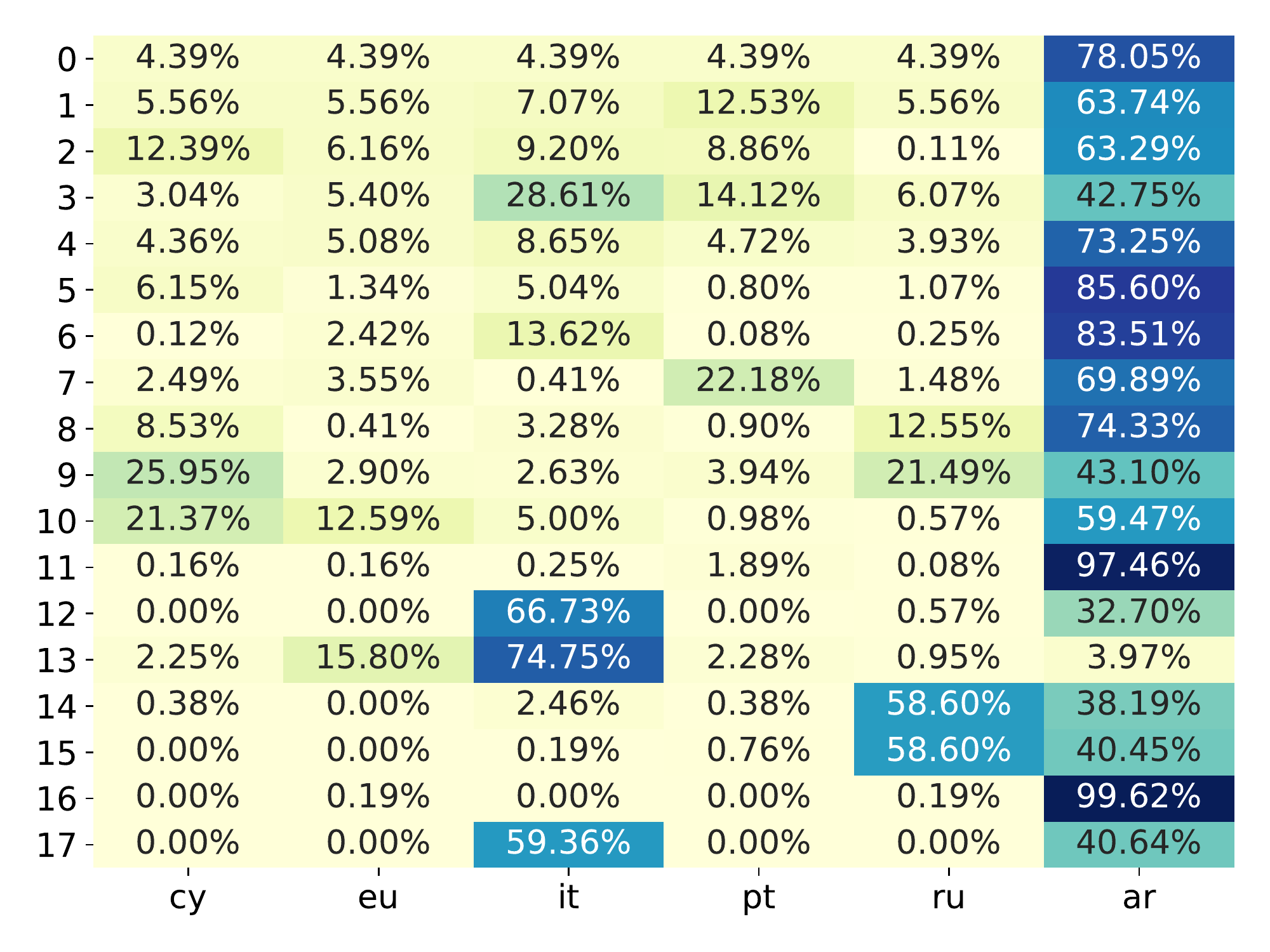}}
        \subfigure[target adapter + guide loss (ar)]{\includegraphics[width=.24\textwidth]{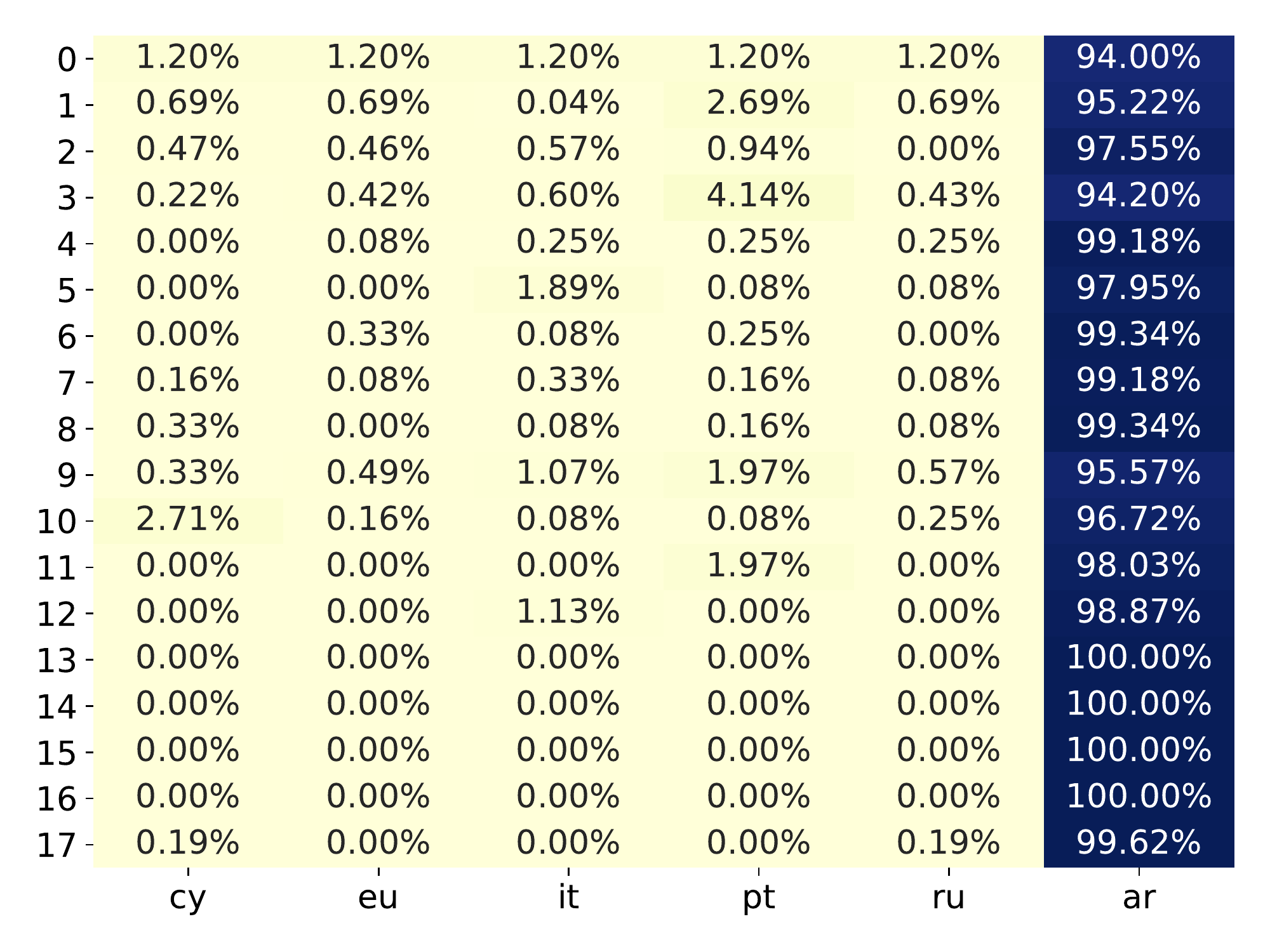}}
        \subfigure[meta-adapter + guide loss (ar)]{\includegraphics[width=.24\textwidth]{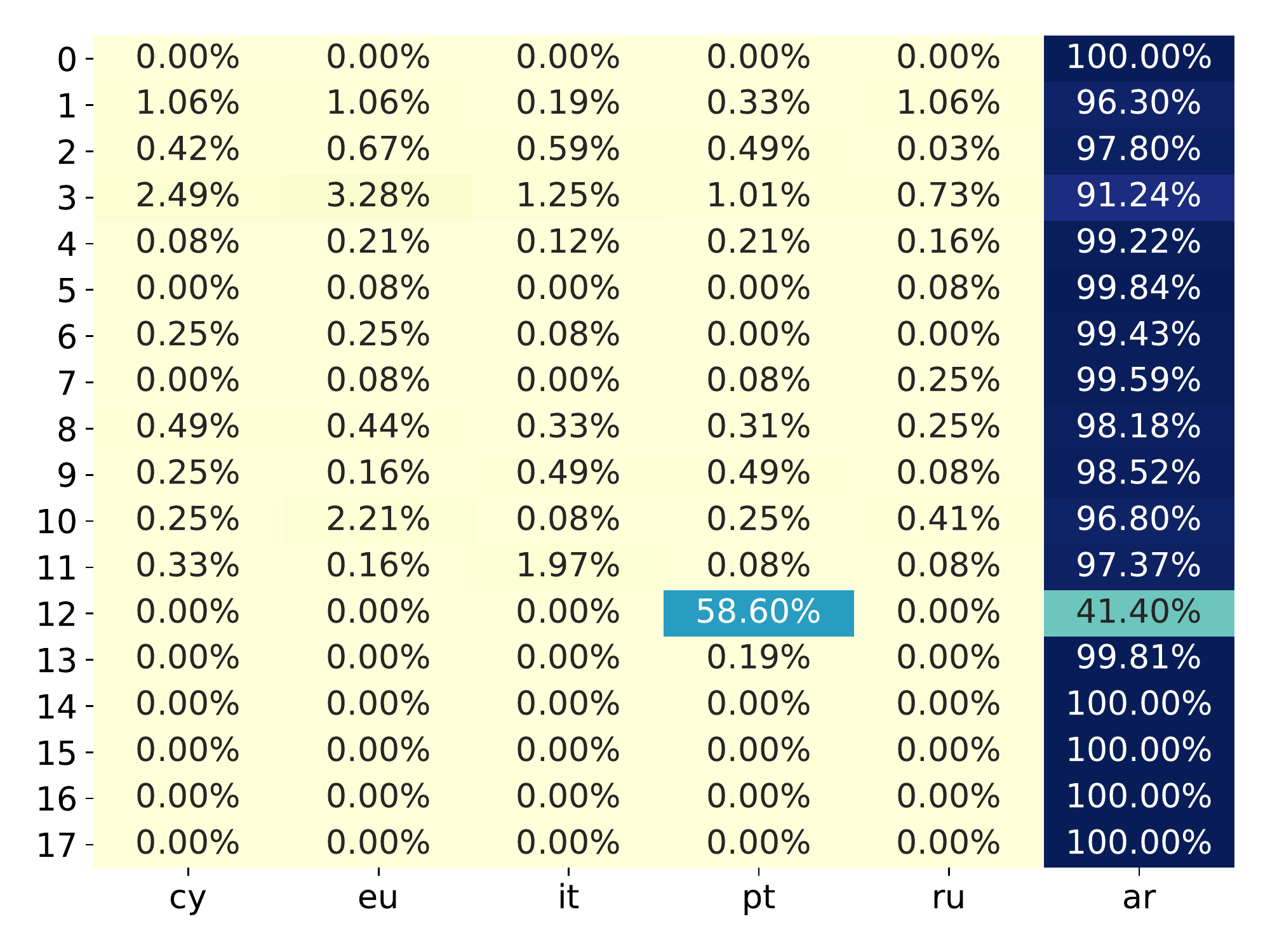}}
        
        %br
        \subfigure[w/o target adapter (br)]{\includegraphics[width=.24\textwidth]{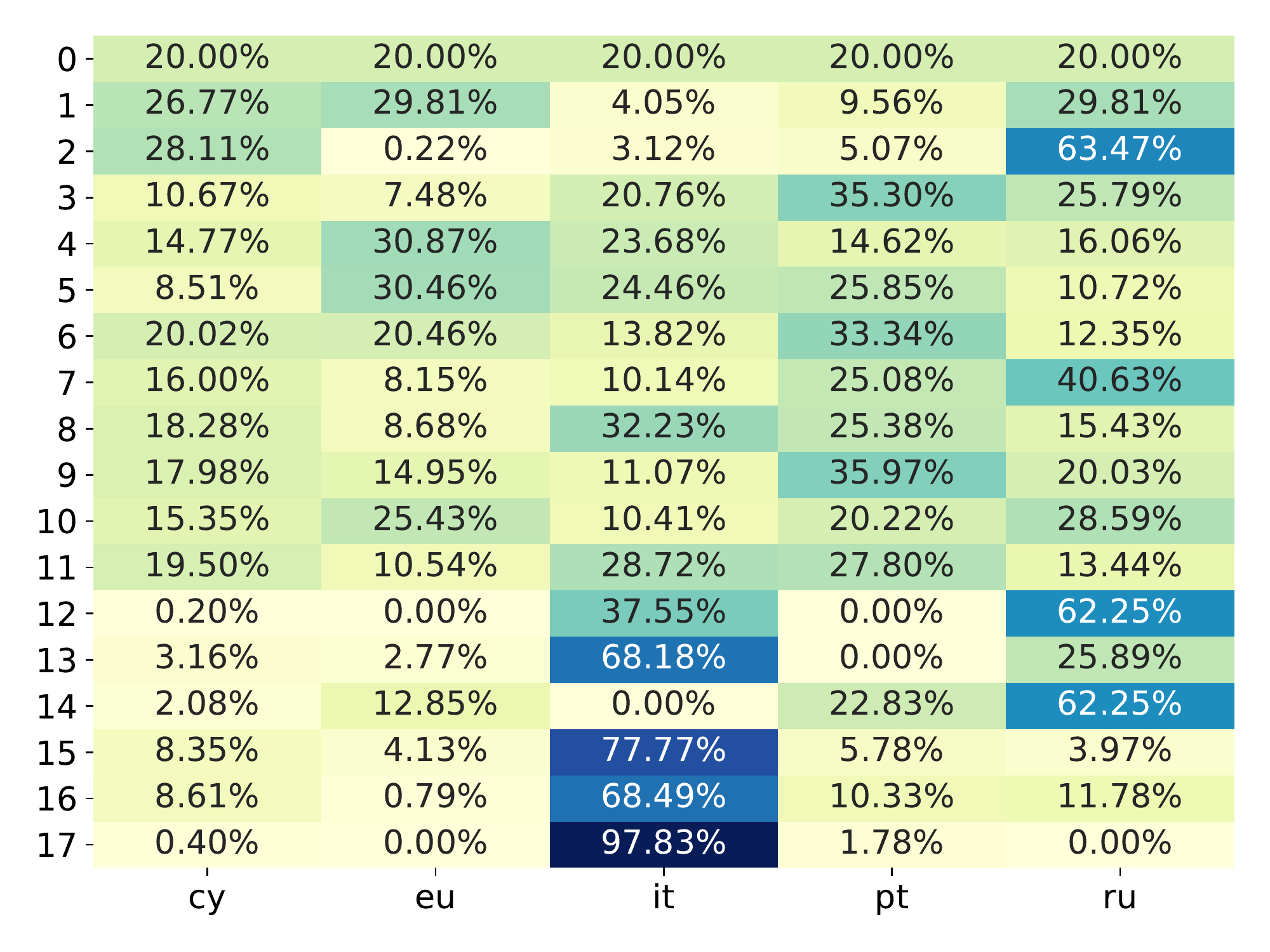}}
        \subfigure[w/o guide loss (br)]{\includegraphics[width=.24\textwidth]{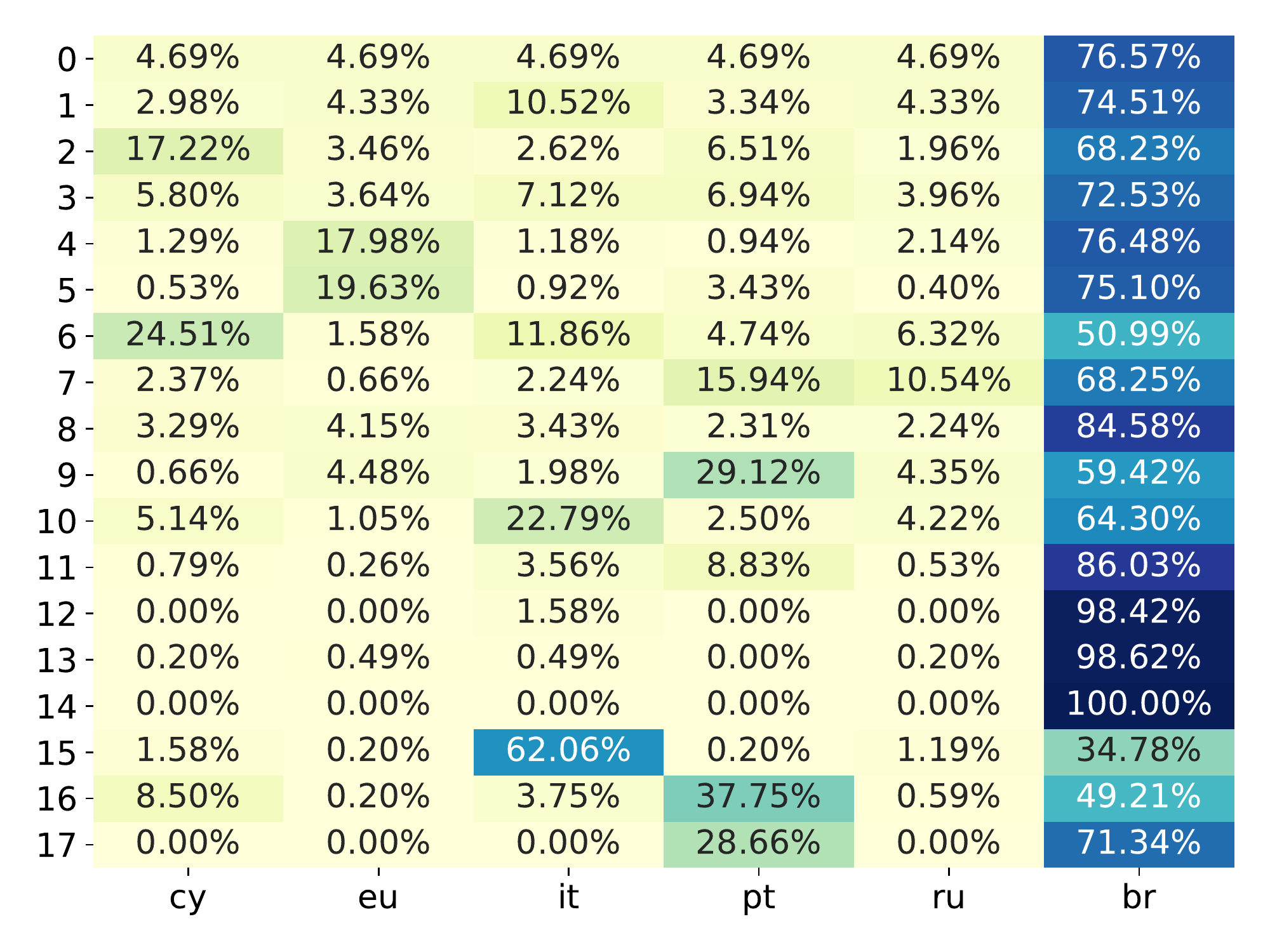}}
        \subfigure[target adapter + guide loss (br)]{\includegraphics[width=.24\textwidth]{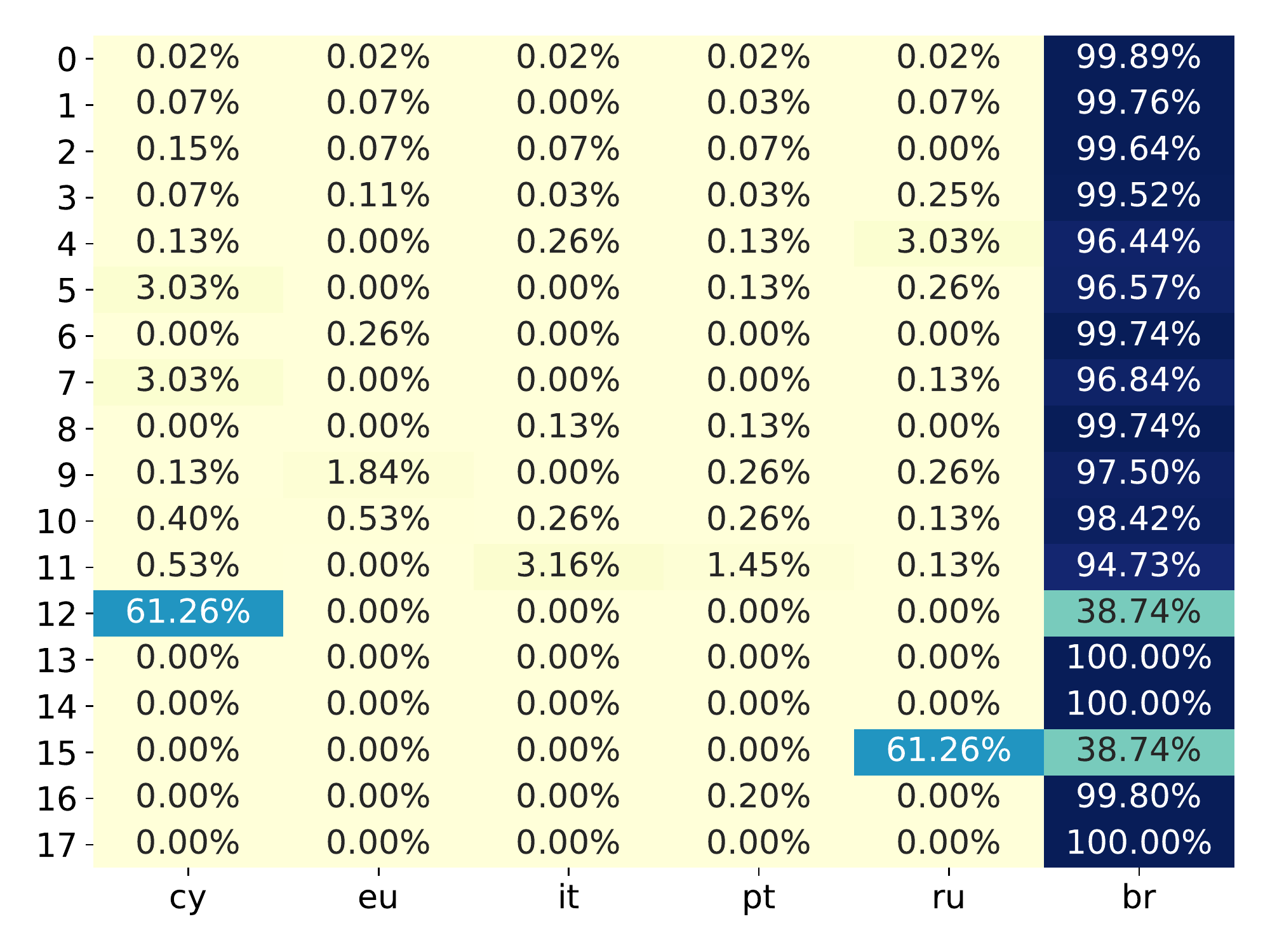}}
        \subfigure[meta-adapter + guide loss (br)]{\includegraphics[width=.24\textwidth]{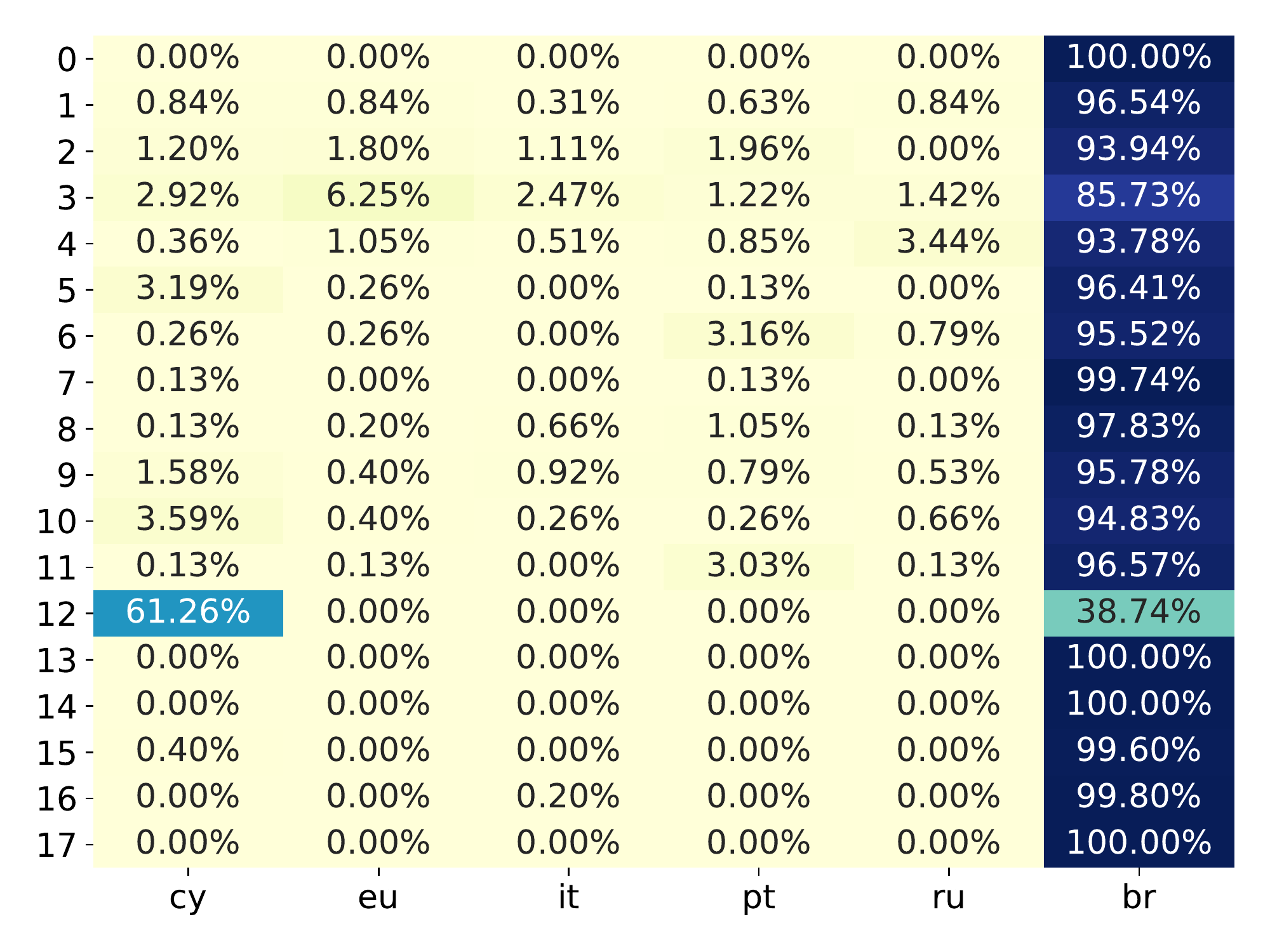}}
        
        %cs
        \subfigure[w/o target adapter (cs)]{\includegraphics[width=.24\textwidth]{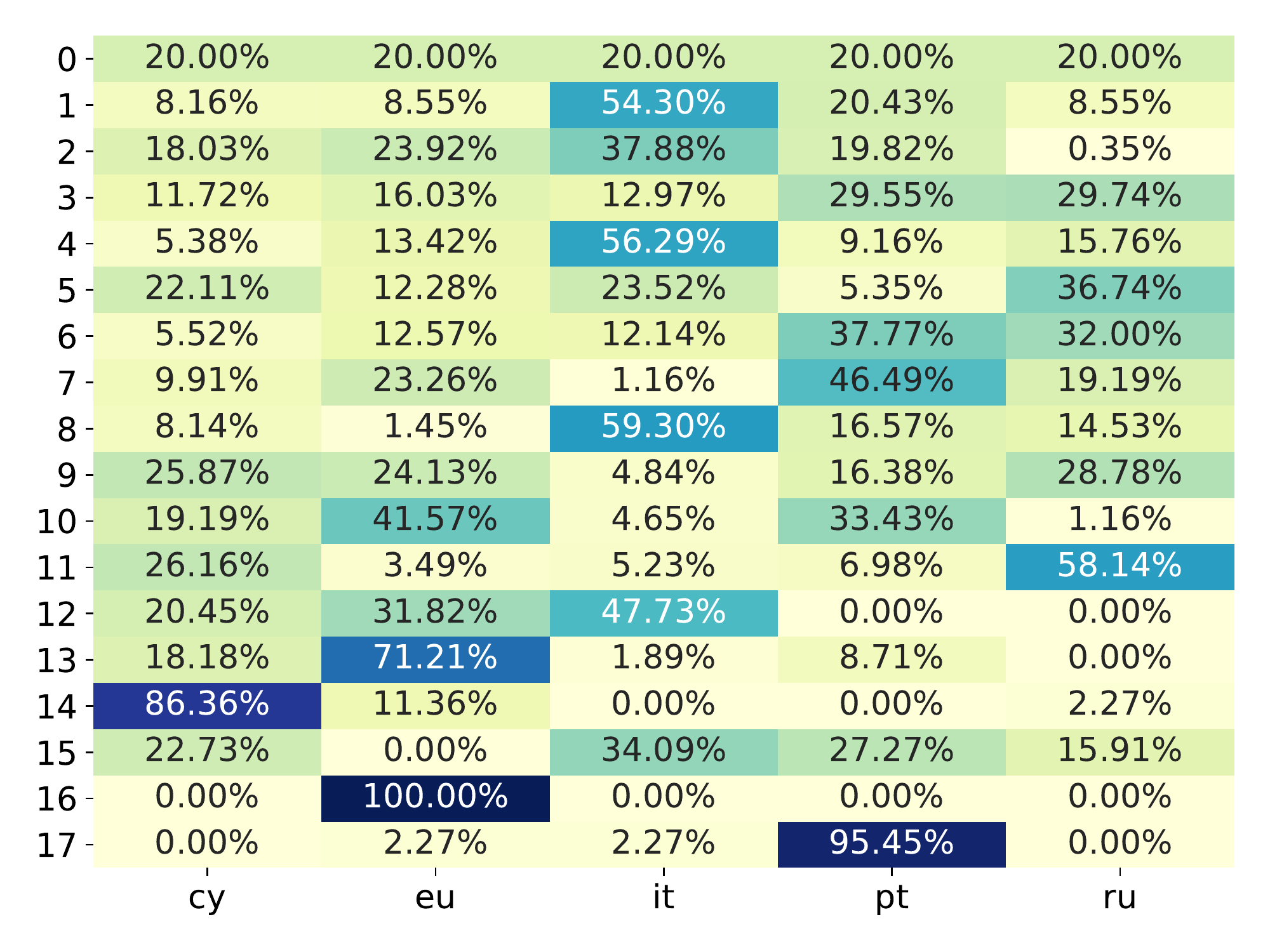}}
        \subfigure[w/o guide loss (cs)]{\includegraphics[width=.24\textwidth]{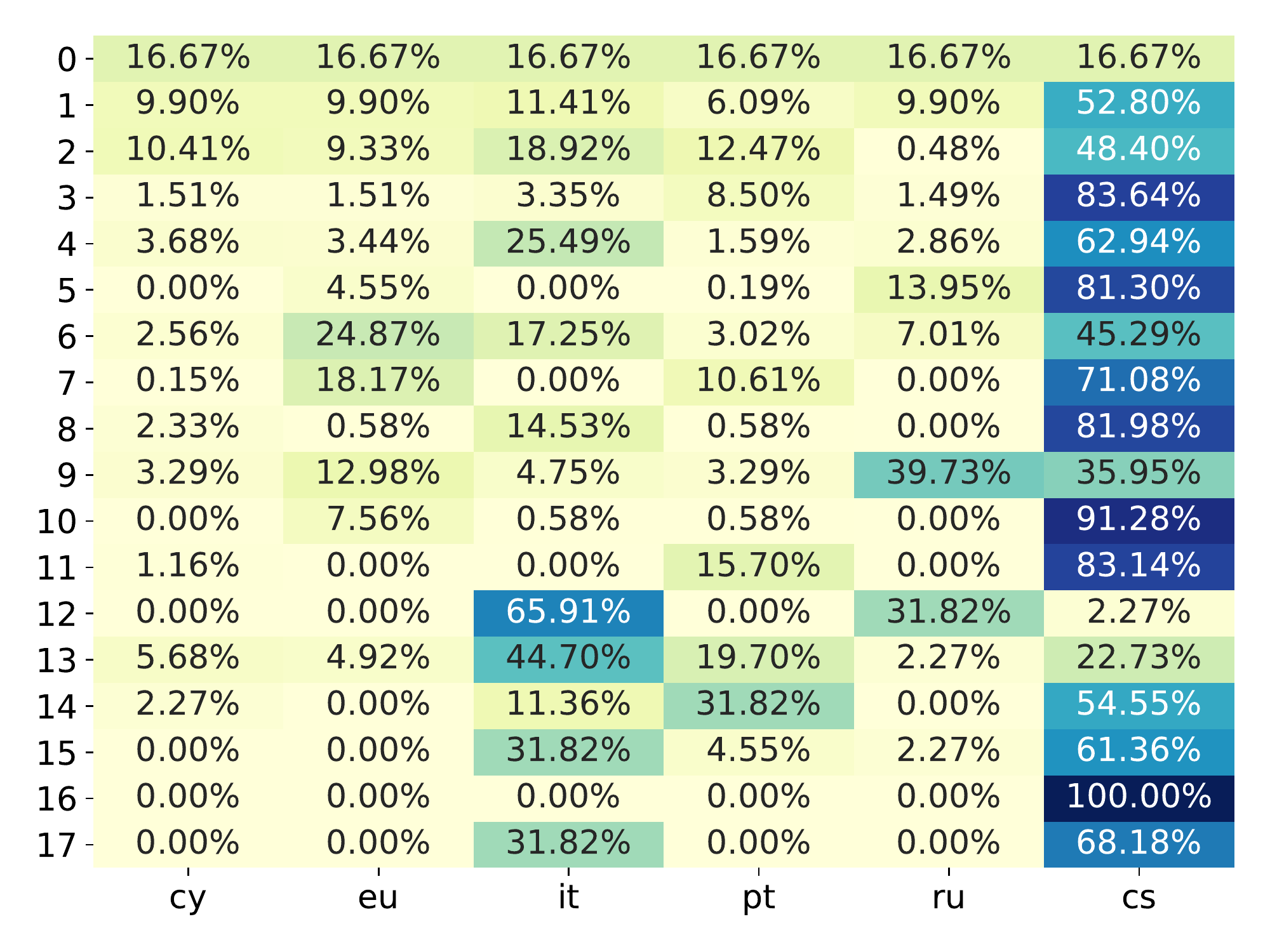}}
        \subfigure[target adapter + guide loss (cs)]{\includegraphics[width=.24\textwidth]{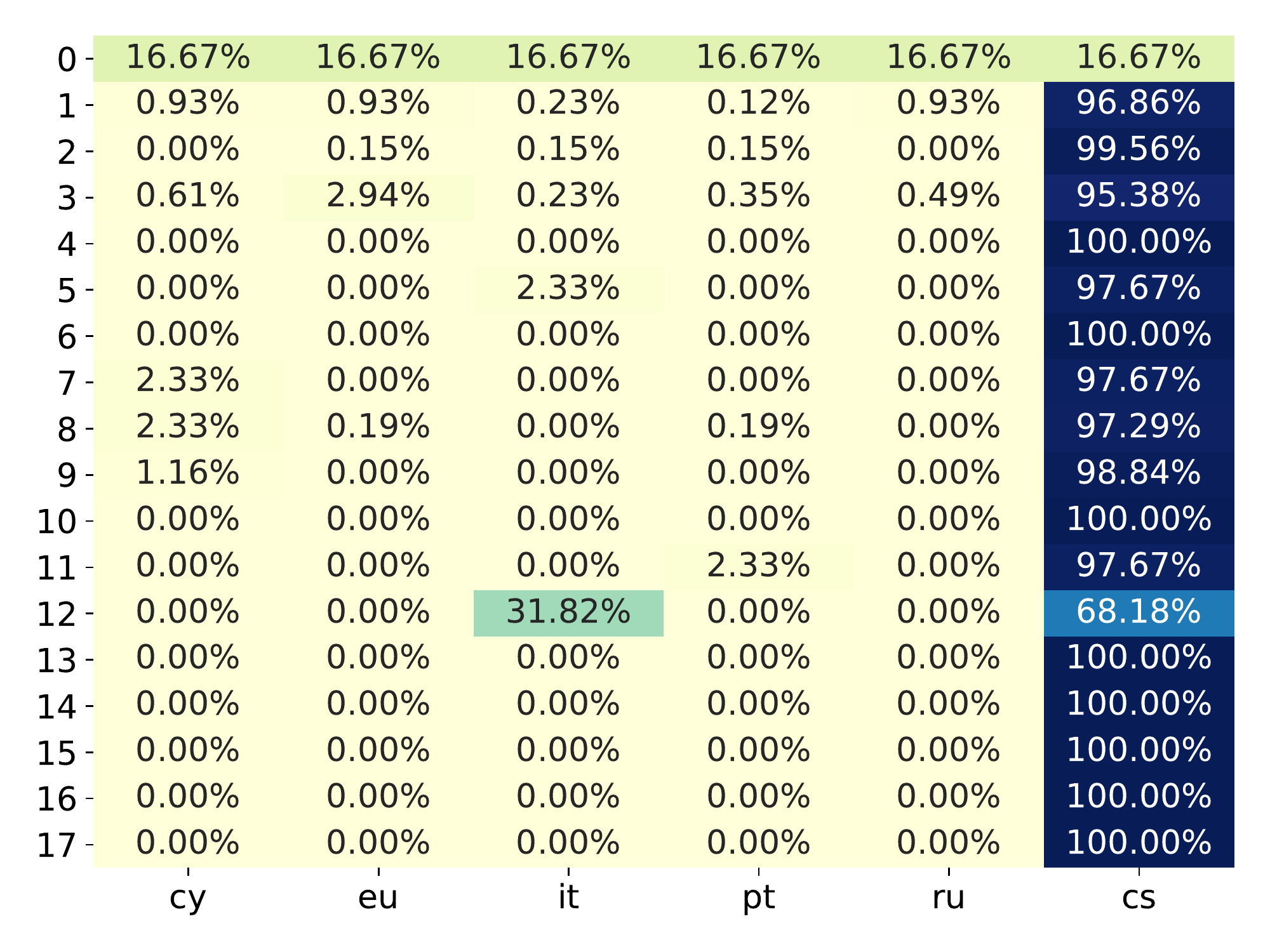}}
        \subfigure[meta-adapter + guide loss (cs)]{\includegraphics[width=.24\textwidth]{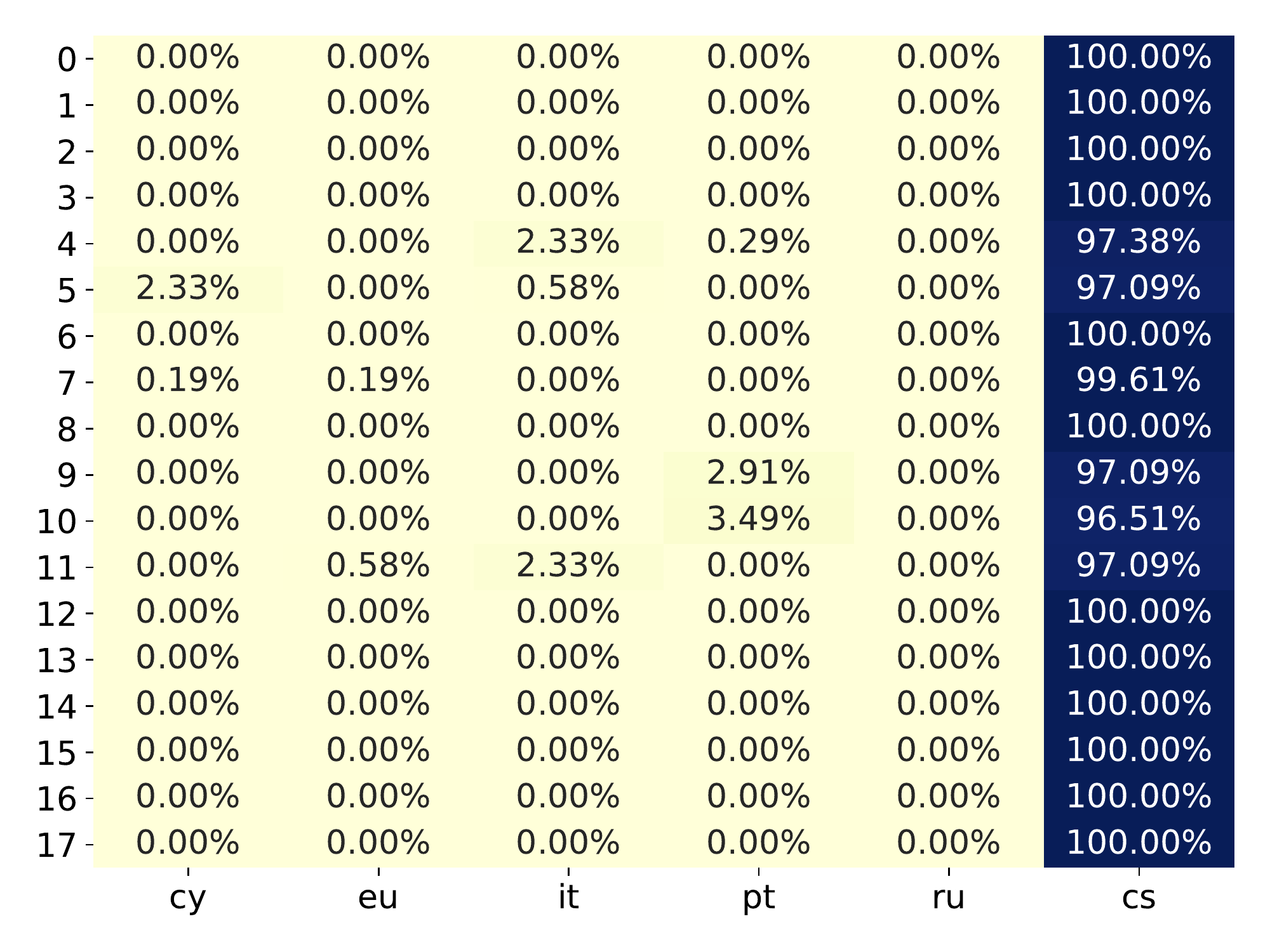}}
        
        %ro
        \subfigure[w/o target adapter (ro)]{\includegraphics[width=.24\textwidth]{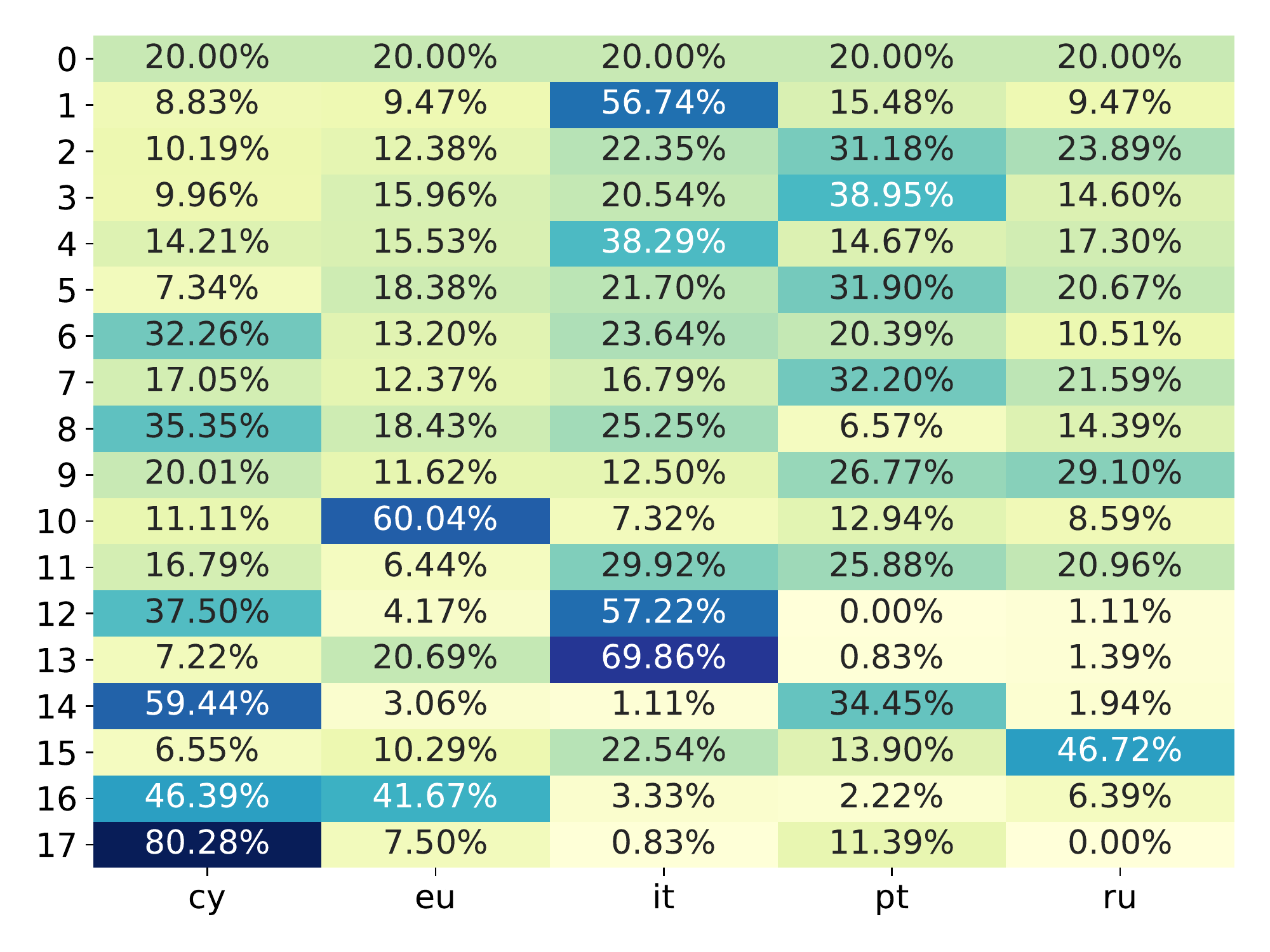}}
        \subfigure[w/o guide loss (ro)]{\includegraphics[width=.24\textwidth]{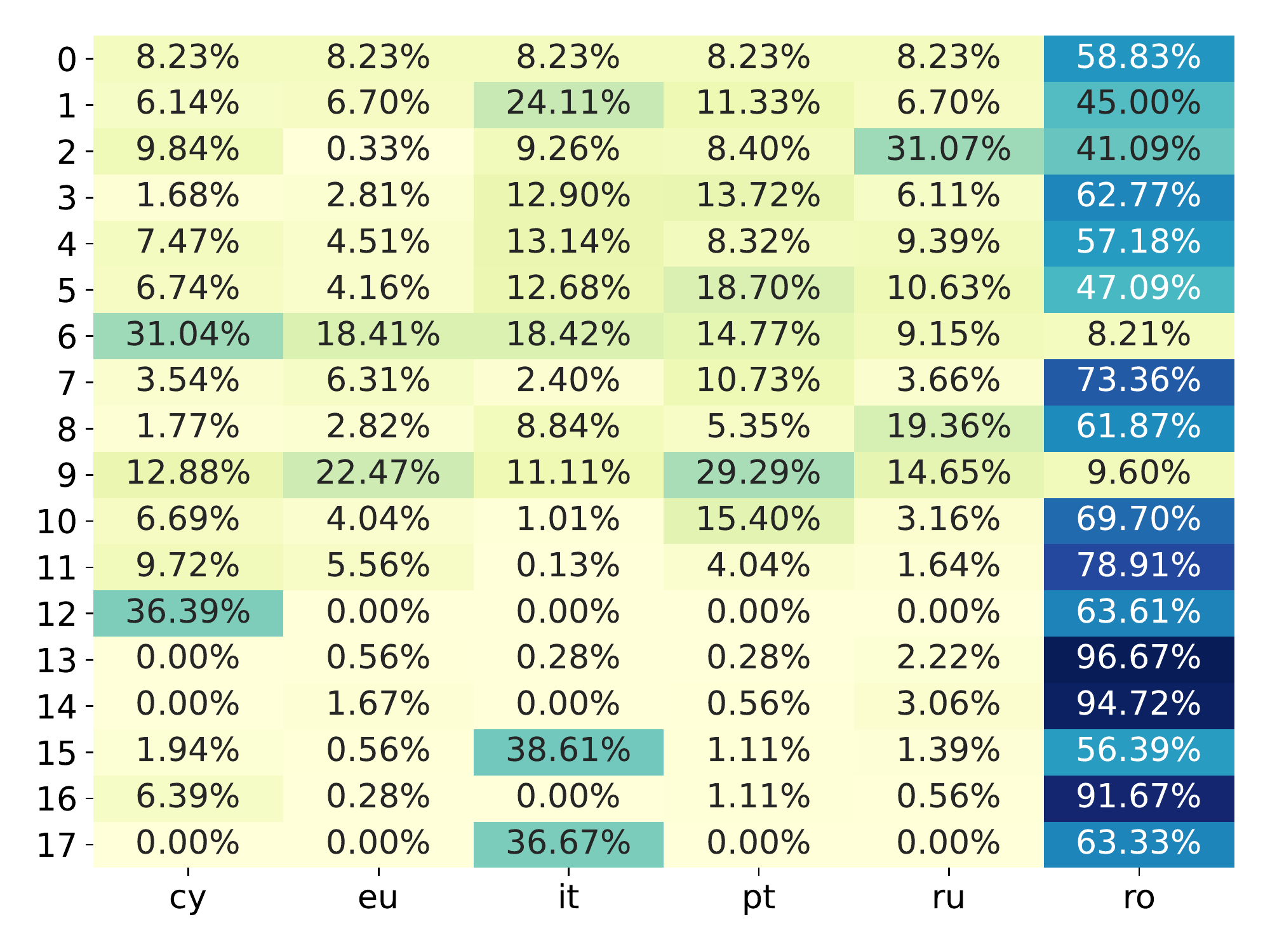}}
        \subfigure[target adapter + guide loss (ro)]{\includegraphics[width=.24\textwidth]{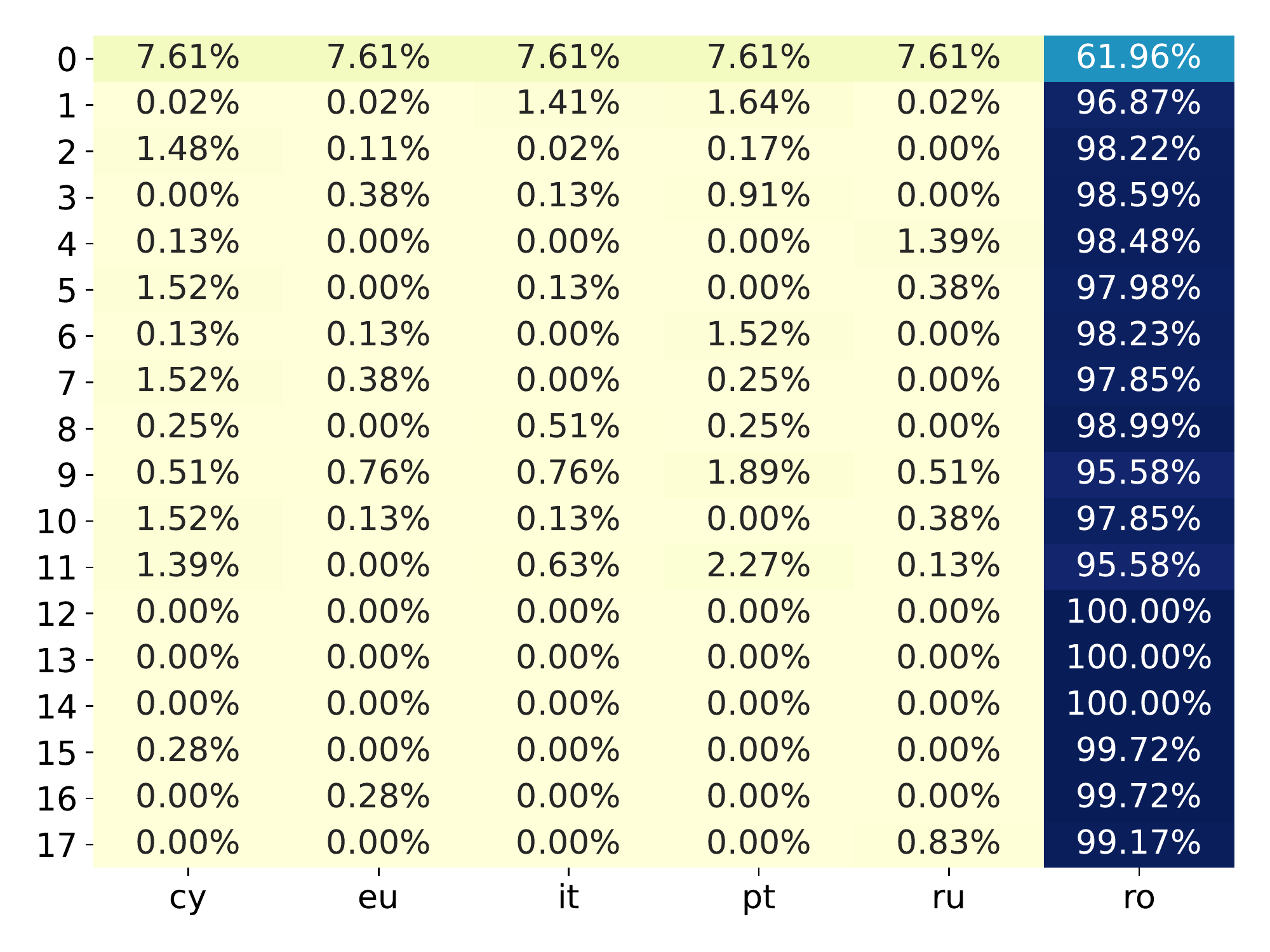}}
        \subfigure[meta-adapter + guide loss (ro)]{\includegraphics[width=.24\textwidth]{figures/attn_figures/meta_guide1_ar-eps-converted-to.pdf}}

        \caption{Attention matrices of five low-resource target languages. A row in the figure denotes a language, whose four settings are: (1) without target adapter, (2) with target adapter but no guide loss ($\gamma=0$), (3) with target adapter and guide loss, and (4) \adaptall.  Column index indicates the Transformer layer number, where 0th to 11th layers are encoders, 12th to 17th are decoders. \textit{Best viewed in color and zoomed in.}}
        \label{fig-attention-otherfour}
    \end{minipage}
    \end{figure*}

\subsubsection{Impact of pre-training epochs for \metaad}
To validate the meta-training effects for the~\metaad, we select checkpoints of 5 pre-trained epochs $\{10, 15, 20, 25, 30\}$ and fine-tune them following the same setting as explained in Section~\ref{sec-exp}. We present the results in \figurename~\ref{fig-abla-pretrain}. It could be found that the WERs are reduced with more pre-training epochs, indicating the effectiveness of meta-learning.

For comparison, we also conduct the same experiment on another adapter pre-trained on source languages using the conventional multi-objective learning (MOL) method and visualize the average WERs in \figurename~\ref{fig-pretrain-approach}. It is clear that with the more pre-training epochs, the MOL-trained adapter tends to overfit the source data and performs worse on the target languages.

% \begin{table}[htbp]
% \centering
% \caption{Impact of the pre-training epochs for \metaad}
% \label{tab-metapretrain}
% \begin{tabular}{ccccccc}
% \toprule
% Target & 10     & 15     & 20     & 25    & 30     \\
% \hline
% ro  & 46.47  & 46.39  & 46.05  & 45.81 & \textbf{44.59}  \\
% cs     & 37.59  & 38.17  & 37.81  & 37.68 & \textbf{37.13}  \\
% br    & 60.07  & 60.82  & 60.33  & 59.20  & \textbf{58.47}  \\
% ar    & 47.67  & 47.86  & 47.30   & 48.39 & \textbf{46.82}  \\
% uk & 52.31  & 50.47  & 50.50   & \textbf{49.27} & 49.36  \\
% \hline
% Average      & 48.82 & 48.74 & 48.40 & 48.07 & \textbf{47.27} \\
% \bottomrule
% \end{tabular}
% \end{table}

\subsubsection{Analyzing the weight of guide loss for \adaptfu}
\begin{figure*}[t!]
    \centering
    \subfigure[Pre-training of \metaad]{
    \includegraphics[width=.45\textwidth]{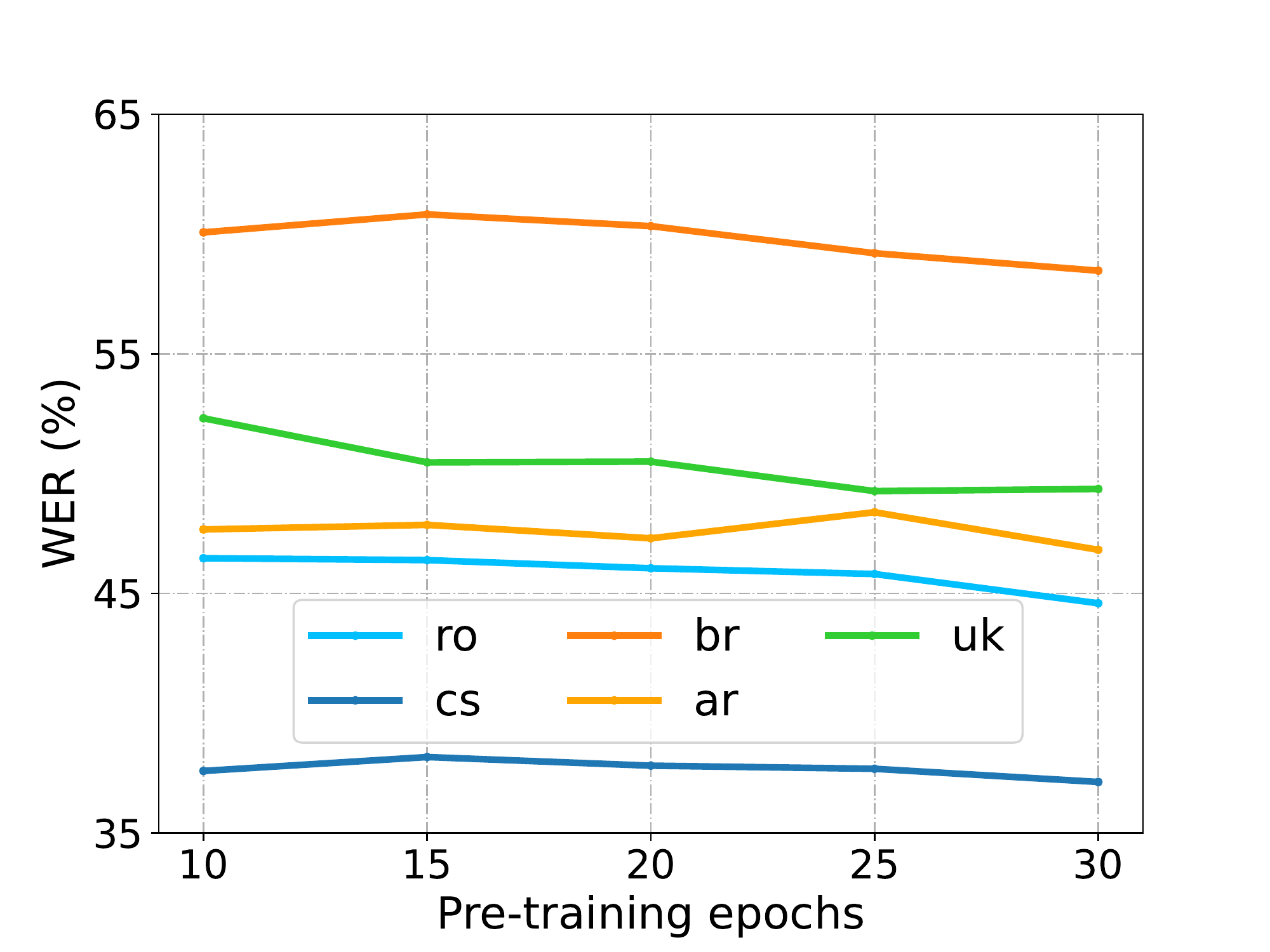}
    \label{fig-abla-pretrain}
    }
    \subfigure[Guide loss of \adaptfu]{
    \includegraphics[width=.45\textwidth]{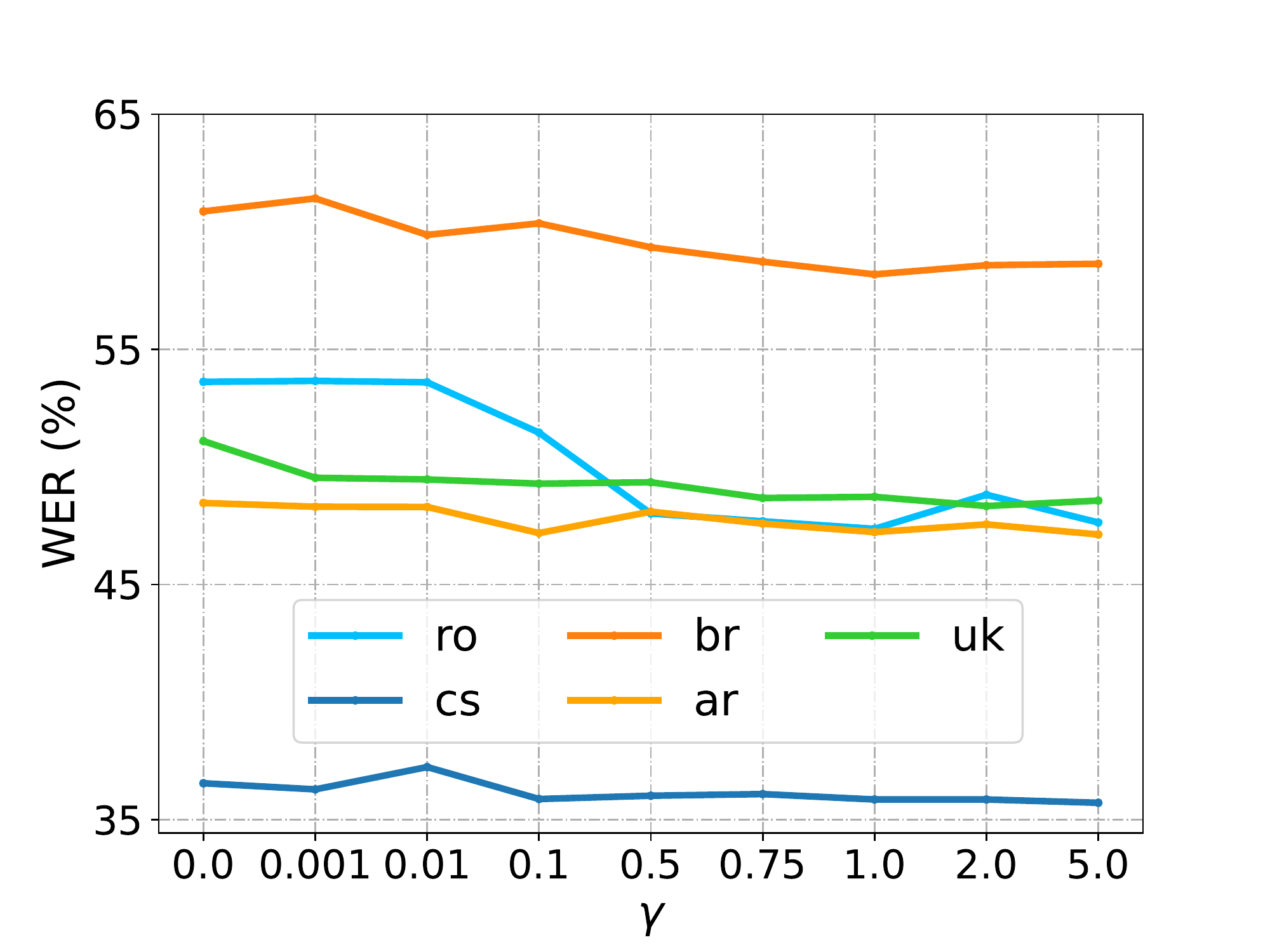}
    \label{fig-abla-guide}
    }
    \caption{Analysis of (a) pre-training epochs of \metaad and (b) importance of guide loss in \adaptfu.}
    \label{fig-ablation}
\end{figure*}

We then analyze the impacts of the weight $\gamma$ of the proposed guide loss within $\{0, 0.001, 0.01, 0.1, 0.5, 0.75, 1.0\}$ for the~\adaptfu. As shown in \figurename~\ref{fig-abla-guide}, the model performances on the 5 languages generally get improved with the increasing of $\gamma$ when $\gamma < 0.5$. When $\gamma \geq 0.5$, the WER may vary among languages. The best overall performance is obtained when $\gamma = 1$.
In real applications, the value of $\gamma$ needs to tune on the target dataset.
%increasing $\gamma$ generally benefits the improvement of average performance, indicating the correctness of our assumption that \adaptfu layers can learn better under the guidance. We can also observe that when with the increasing of $\gamma$, \adaptfu achieves comparable results to full-model fine-tuning at the weight of 0.1 and surpasses it consistently at 0.5, 0.75 and 1.0.

% \begin{figure}[htbp]
%     \centering
%     \includegraphics[width=.4\textwidth]{figures/fig-guideloss.eps}
%     \caption{Results of \adaptfu with different guide loss weight $\gamma$.}
%     \label{fig-guide-loss-weight}
% \end{figure}
% \begin{table}[tb]
% \centering
% \caption{Guide loss weight}
% \label{tab-guide-loss-weight}
% \begin{tabular}{ccccccc}
% \toprule
% Target       & 0      & 0.001 & 0.01   & 0.1    & 0.5    & 1      \\
% \hline
% ro  & 53.62  & 53.66 & 53.60  & 51.46  & 48.02  & 47.37  \\
% cs     & 36.55  & 36.29 & 37.24  & 35.88  & 36.02  & 35.86  \\
% br    & 60.87  & 61.42 & 59.87  & 60.36  & 59.34  & 58.19  \\
% ar    & 48.47  & 48.31 & 48.30   & 47.19  & 48.10   & 47.23  \\
% uk & 51.10   &   49.54    & 49.47  & 49.29  & 49.35  & 48.73  \\
% \hline
% Average      & 50.12 &  49.84 & 49.70 & 48.84 & 48.17 & 47.48 \\
% \bottomrule
% \end{tabular}
% \end{table}

\subsubsection{How much information can be shared across languages}
Although \adaptfu improves the WER results, we do not know whether and how much it could benefit from other languages. Therefore, we conduct two experiments to validate this. Firstly, we examine how much the other languages can contribute without using the adapters from target languages to see whether additional gains can be obtained with only source adapters. \tablename~\ref{tab-srconly} shows the results. It can be found that even without the target adapter, \adaptfu can still improve the performance for most of the languages except for Romanian, indicating the effectiveness of learning language information from source adapters.
\begin{table}[htbp]
\centering
\caption{WER results of \adaptfu with or without Adapter ${L_T}$. Fusion guide loss is set to 0 for \adaptfu with Adapter ${L_T}$.}
\label{tab-srconly}
\begin{tabular}{ccccc}
\toprule
Target      & Head   & w/o Adapter ${L_T}$ & w/ Adapter ${L_T}$ \\
\hline
ro  & 63.98  & 67.83    & 53.62     \\
cs     & 75.12  & 55.06    & 36.55  \\
br    & 82.80   & 77.04    & 60.87  \\
ar    & 81.70   & 64.68    & 48.47  \\
uk & 82.71  & 69.09    & 51.10      \\
\hline
Average      & 77.26 & 66.74    & 50.12   \\
+Weighted & 77.54 & 65.33 & 48.39 \\
\bottomrule
\end{tabular}
\end{table}

In the second experiment, we train two different \adaptfu models on Ukrainian by fusing the target-language adapter and one source-language adapter to analyze the contributions of different source languages. 
Specifically, we choose Italian and Russian as the source languages.
Since Russian is more similar to Ukrainian than Italian, we expect more gains of \adaptfu trained with the Russian adapter.
% The results are presented in \tablename~\ref{tab-exp-on-uk}
The results align with our expectation.
We observe that the \adaptfu with Italian adapter obtains a WER of 48.70, while with the Russian adapter, the WER is 47.73, indicating that \adaptfu could transfer more useful knowledge from Russian than Italian to model the Ukrainian language.
% \begin{table}[h]
%     \centering
%     \caption{Comparison between different source languages}
%     \label{tab-exp-on-uk}
%     \begin{tabular}{c|ccc}
%     \toprule
%       Target  & Adapter & w/ Italian  & w/ Russian \\
%       \hline
%         Ukrainian & 50.84 & 48.70 & 47.73 \\
%     \bottomrule
%     \end{tabular}
% \end{table}

\subsection{Attention Visualization}
To further show the relationship between source and target languages, we visualize the attention maps for each target language.
The attention value reflects their similarities.
\figurename~\ref{fig-attention-otherfour} shows the results of three different types of languages: (1) without target adapter, (2) with target adapter but no guide loss ($\gamma=0$), (3) with target adapter and guide loss, and (4) with target \metaad and guide loss.

We take the Ukrainian (uk) as an example.
Firstly, from the figure on the left, we can observe a trend that \adaptfu layers tend to pay more attention to the Russian (ru)'s adapter, which could be because of the linguistic similarity between Ukrainian and Russian. However, after introducing the target adapter, \adaptfu layers obviously turn to focus more on the target adapter, but there are still diverse attentions across other languages. By introducing the guide loss, the \adaptfu layers are forced to pay more attention to the target adapter and fusing less information from other languages.

We also notice that in the first encoder layer, the attention distribution seems to be uniform across the source languages. By analyzing the outputs, we found that the adapters in the first layer tend to keep the backbone representation unchanged via the residual connection. The same phenomenon can also be observed in the Czech (cs) target language. A possible reason could be that the first layer is to extract general acoustic features which are language-independent. Since we observe a similar trend in the first decoder layer (layer 12) that the attention distributions tend to be more distracted, we thus assume that adapters in the bottom layers in both the encoder and decoder are less important for cross-lingual adaptation, which we conduct experiments in next subsection to analyze the performance of fusing different adapters.

% \begin{figure*}[t!]
%     \centering
%     \subfigure[w/o target adapter]{\includegraphics[width=.22\textwidth]{figures/attn_figures/srconly_uk.eps}}
%     \subfigure[w/o guide loss]{\includegraphics[width=.22\textwidth]{figures/attn_figures/guide0_uk.eps}}
%     \subfigure[target adapter + guide loss]{\includegraphics[width=.22\textwidth]{figures/attn_figures/guide1_uk.eps}}
%     \caption{Attention matrices under 3 different settings: (a) without target adapter, (b) with target adapter but no guide loss ($\gamma=0$), and (3) with target adapter and guide loss. Column index indicates the Transformer layer number, where 0th to 11th layers are encoders, 12th to 17th are decoders.}
%     \label{fig-uk}
% \end{figure*}
\subsection{Do all Adapter layers need to be fused?}
By observing the attention maps, we notice that for some layers, the attention seems to focus solely on the target adapter with a $100\%$ attention score. This phenomenon occurs more frequently in the higher decoder layers, i.e., 12th to 17th layers in Fig.~\ref{fig-attention-otherfour}. In such cases, the fusion seems not to be necessary. We doubt whether we can achieve comparable performance while fusing adapters in part of the layers only. Therefore, we conduct the ablation experiments by only fusing part of the layers. The results are presented in~\tablename~\ref{tab-enc-dec}. Although some languages (e.g., Breton) can retain the performance by only fusing 2 bottom layers, fusing more layers generally lead to better performance.

% a possible reason could be due to the dynamic value adjustment.

\begin{table}[htbb]
\centering
\caption{Ablation study of the encoder and decoders}
\label{tab-enc-dec}
\begin{tabular}{cccc}
\toprule
Target       & Enc1-Dec1 & Enc12-Dec1 & Enc12-Dec6 \\
\hline
ro  & 48.39     & 48.25      & \textbf{47.37}       \\
cs     & 37.31     & 36.30      & \textbf{35.86}       \\
br    & \textbf{57.85}     & 59.08      & 58.19       \\
ar    & 47.48     & 47.34      & \textbf{47.23}       \\
uk & 50.58     & 48.98      & \textbf{48.73}       \\
\hline
Average      & 48.32    & 47.99       & \textbf{47.48}       \\
+Weighted & 47.04 & 46.55 & \textbf{46.08} \\
\bottomrule
\end{tabular}
\end{table}

\subsection{Training and inference time}
Finally, we compare the average training time of full-model fine-tuning, \metaad and \adaptfu methods per iteration as well as their inference real-time factor (RTF) on the 5 target languages. The RTF metric is used to evaluate the decoding time cost by computing the ratio of the model decoding time to the total utterance duration on the test data. The training and decoding are conducted on 1 GeForce RTX 2080 Ti GPU with batch size 64. The results are shown in \tablename~\ref{tab-time}.

It could be found that the \metaad module significantly accelerates the training process while the \adaptfu introduces minor additional time cost compared with full-model fine-tuning. The RTFs of Full-FT and \metaad are at the same level. The reason that \metaad has slightly lower RTF could be due to its shorter average prediction lengths. On the other hand, the relative RTF increases of 22.12\% brought by \adaptfu is also acceptable.

\begin{table}[tb]
\centering
\caption{Average Training / inference time.}
\label{tab-time}
\begin{tabular}{lll}
\toprule
              & Training Time (sec.)  & RTF \\
\hline
Full-FT    & 0.253 (-)       &        0.045  (-)     \\
\metaad       & 0.143 (43.48\%$\downarrow$)         &  0.043 (4.06\%$\downarrow$)             \\
\adaptfu & 0.263 (3.95\%$\uparrow$)        &     0.055 (22.12\%$\uparrow$)     \\
\bottomrule
\end{tabular}
\end{table}

\section{Conclusions and Future Work}
\label{sec-con}
In this paper, we propose to exploit \metaad and \adaptfu for adapter-based cross-lingual speech recognition.
The proposed \adaptfu leverages the attention mechanism to learn the similarities between the source and target languages during fine-tuning using the adapters.
We also show that the two algorithms can be integrated for better performance.
Experiments on five low-resource languages from Common Voice dataset demonstrate the superiority of the two algorithms. 
In the future, we plan to extend these algorithms to other language families and further improve the training and inference speed of our methods.

% % use section* for acknowledgment
% \section*{Acknowledgment}

\ifCLASSOPTIONcaptionsoff
  \newpage
\fi

\bibliographystyle{IEEEtran}
\bibliography{mybib}

\begin{thebibliography}{10}
\providecommand{\url}[1]{#1}
\csname url@samestyle\endcsname
\providecommand{\newblock}{\relax}
\providecommand{\bibinfo}[2]{#2}
\providecommand{\BIBentrySTDinterwordspacing}{\spaceskip=0pt\relax}
\providecommand{\BIBentryALTinterwordstretchfactor}{4}
\providecommand{\BIBentryALTinterwordspacing}{\spaceskip=\fontdimen2\font plus
\BIBentryALTinterwordstretchfactor\fontdimen3\font minus
  \fontdimen4\font\relax}
\providecommand{\BIBforeignlanguage}[2]{{%
\expandafter\ifx\csname l@#1\endcsname\relax
\typeout{** WARNING: IEEEtran.bst: No hyphenation pattern has been}%
\typeout{** loaded for the language `#1'. Using the pattern for}%
\typeout{** the default language instead.}%
\else
\language=\csname l@#1\endcsname
\fi
#2}}
\providecommand{\BIBdecl}{\relax}
\BIBdecl

\bibitem{wang2019overview}
D.~Wang, X.~Wang, and S.~Lv, ``An overview of end-to-end automatic speech
  recognition,'' \emph{Symmetry}, vol.~11, no.~8, p. 1018, 2019.

\bibitem{karita2019comparative}
S.~Karita, N.~Chen, T.~Hayashi, T.~Hori, H.~Inaguma, Z.~Jiang, M.~Someki,
  N.~E.~Y. Soplin, R.~Yamamoto, X.~Wang \emph{et~al.}, ``A comparative study on
  transformer vs rnn in speech applications,'' in \emph{2019 IEEE Automatic
  Speech Recognition and Understanding Workshop (ASRU)}.\hskip 1em plus 0.5em
  minus 0.4em\relax IEEE, 2019, pp. 449--456.

\bibitem{pratap2020massively}
V.~Pratap, A.~Sriram, P.~Tomasello, A.~Hannun, V.~Liptchinsky, G.~Synnaeve, and
  R.~Collobert, ``Massively multilingual asr: 50 languages, 1 model, 1 billion
  parameters,'' \emph{Proc. Interspeech 2020}, pp. 4751--4755, 2020.

\bibitem{Hou2020}
\BIBentryALTinterwordspacing
W.~Hou, Y.~Dong, B.~Zhuang, L.~Yang, J.~Shi, and T.~Shinozaki, ``{Large-Scale
  End-to-End Multilingual Speech Recognition and Language Identification with
  Multi-Task Learning},'' in \emph{Proc. Interspeech 2020}, 2020, pp.
  1037--1041. [Online]. Available:
  \url{http://dx.doi.org/10.21437/Interspeech.2020-2164}
\BIBentrySTDinterwordspacing

\bibitem{hou21b_interspeech}
W.~Hou, J.~Wang, X.~Tan, T.~Qin, and T.~Shinozaki, ``{Cross-Domain Speech
  Recognition with Unsupervised Character-Level Distribution Matching},'' in
  \emph{Proc. Interspeech 2021}, 2021, pp. 3425--3429.

\bibitem{conneau2020unsupervised}
A.~Conneau, A.~Baevski, R.~Collobert, A.~Mohamed, and M.~Auli, ``{Unsupervised
  Cross-Lingual Representation Learning for Speech Recognition},'' in
  \emph{Proc. Interspeech 2021}, 2021, pp. 2426--2430.

\bibitem{chuangsuwanich2016multilingual}
E.~Chuangsuwanich, ``Multilingual techniques for low resource automatic speech
  recognition,'' Massachusetts Institute of Technology Cambridge United States,
  Tech. Rep., 2016.

\bibitem{adams2019massively}
O.~Adams, M.~Wiesner, S.~Watanabe, and D.~Yarowsky, ``Massively multilingual
  adversarial speech recognition,'' in \emph{Proceedings of the 2019 Conference
  of the North American Chapter of the Association for Computational
  Linguistics: Human Language Technologies, Volume 1 (Long and Short Papers)},
  2019, pp. 96--108.

\bibitem{zhou2018multilingual}
S.~Zhou, S.~Xu, and B.~Xu, ``Multilingual end-to-end speech recognition with a
  single transformer on low-resource languages,'' \emph{arXiv preprint
  arXiv:1806.05059}, 2018.

\bibitem{hsu2020meta}
J.-Y. Hsu, Y.-J. Chen, and H.-y. Lee, ``Meta learning for end-to-end
  low-resource speech recognition,'' in \emph{ICASSP 2020-2020 IEEE
  International Conference on Acoustics, Speech and Signal Processing
  (ICASSP)}.\hskip 1em plus 0.5em minus 0.4em\relax IEEE, 2020, pp. 7844--7848.

\bibitem{hou2021meta}
W.~{Hou}, Y.~{Wang}, S.~{Gao}, and T.~{Shinozaki}, ``Meta-adapter: Efficient
  cross-lingual adaptation with meta-learning,'' in \emph{ICASSP 2021 - 2021
  IEEE International Conference on Acoustics, Speech and Signal Processing
  (ICASSP)}, 2021.

\bibitem{kannan2019large}
A.~Kannan, A.~Datta, T.~N. Sainath, E.~Weinstein, B.~Ramabhadran, Y.~Wu,
  A.~Bapna, Z.~Chen, and S.~Lee, ``{Large-Scale Multilingual Speech Recognition
  with a Streaming End-to-End Model},'' in \emph{Proc. Interspeech 2019}, 2019,
  pp. 2130--2134.

\bibitem{winata2020adapt}
G.~I. Winata, G.~Wang, C.~Xiong, and S.~Hoi, ``{Adapt-and-Adjust: Overcoming
  the Long-Tail Problem of Multilingual Speech Recognition},'' in \emph{Proc.
  Interspeech 2021}, 2021, pp. 2451--2455.

\bibitem{pfeiffer2020adapterfusion}
\BIBentryALTinterwordspacing
J.~Pfeiffer, A.~Kamath, A.~R{\"u}ckl{\'e}, K.~Cho, and I.~Gurevych,
  ``{A}dapter{F}usion: Non-destructive task composition for transfer
  learning,'' in \emph{Proceedings of the 16th Conference of the European
  Chapter of the Association for Computational Linguistics: Main Volume}.\hskip
  1em plus 0.5em minus 0.4em\relax Online: Association for Computational
  Linguistics, Apr. 2021, pp. 487--503. [Online]. Available:
  \url{https://aclanthology.org/2021.eacl-main.39}
\BIBentrySTDinterwordspacing

\bibitem{finn2017model}
C.~Finn, P.~Abbeel, and S.~Levine, ``Model-agnostic meta-learning for fast
  adaptation of deep networks,'' in \emph{International Conference on Machine
  Learning}.\hskip 1em plus 0.5em minus 0.4em\relax PMLR, 2017, pp. 1126--1135.

\bibitem{ke2006language}
J.~Ke and J.~H. Holland, ``Language origin from an emergentist perspective,''
  \emph{Applied Linguistics}, vol.~27, no.~4, pp. 691--716, 2006.

\bibitem{macneilage2010origin}
P.~F. MacNeilage, \emph{The origin of speech}.\hskip 1em plus 0.5em minus
  0.4em\relax Oxford University Press, 2010, no.~10.

\bibitem{frayer200014}
D.~W. Frayer and C.~Nicolay, ``14 fossil evidence for the origin of speech
  sounds,'' 2000.

\bibitem{watanabe2017language}
S.~Watanabe, T.~Hori, and J.~R. Hershey, ``Language independent end-to-end
  architecture for joint language identification and speech recognition,'' in
  \emph{2017 IEEE Automatic Speech Recognition and Understanding Workshop
  (ASRU)}.\hskip 1em plus 0.5em minus 0.4em\relax IEEE, 2017, pp. 265--271.

\bibitem{kim2017joint}
S.~Kim, T.~Hori, and S.~Watanabe, ``Joint ctc-attention based end-to-end speech
  recognition using multi-task learning,'' in \emph{2017 IEEE international
  conference on acoustics, speech and signal processing (ICASSP)}.\hskip 1em
  plus 0.5em minus 0.4em\relax IEEE, 2017, pp. 4835--4839.

\bibitem{toshniwal2018multilingual}
S.~Toshniwal, T.~N. Sainath, R.~J. Weiss, B.~Li, P.~Moreno, E.~Weinstein, and
  K.~Rao, ``Multilingual speech recognition with a single end-to-end model,''
  in \emph{2018 IEEE international conference on acoustics, speech and signal
  processing (ICASSP)}.\hskip 1em plus 0.5em minus 0.4em\relax IEEE, 2018, pp.
  4904--4908.

\bibitem{li2019bytes}
B.~Li, Y.~Zhang, T.~Sainath, Y.~Wu, and W.~Chan, ``Bytes are all you need:
  End-to-end multilingual speech recognition and synthesis with bytes,'' in
  \emph{ICASSP 2019-2019 IEEE International Conference on Acoustics, Speech and
  Signal Processing (ICASSP)}.\hskip 1em plus 0.5em minus 0.4em\relax IEEE,
  2019, pp. 5621--5625.

\bibitem{datta2020language}
A.~Datta, B.~Ramabhadran, J.~Emond, A.~Kannan, and B.~Roark,
  ``Language-agnostic multilingual modeling,'' in \emph{ICASSP 2020-2020 IEEE
  International Conference on Acoustics, Speech and Signal Processing
  (ICASSP)}.\hskip 1em plus 0.5em minus 0.4em\relax IEEE, 2020, pp. 8239--8243.

\bibitem{li2021scaling}
B.~Li, R.~Pang, T.~N. Sainath, A.~Gulati, Y.~Zhang, J.~Qin, P.~Haghani, W.~R.
  Huang, and M.~Ma, ``Scaling end-to-end models for large-scale multilingual
  asr,'' in \emph{2021 IEEE Automatic Speech Recognition and Understanding
  Workshop (ASRU)}.\hskip 1em plus 0.5em minus 0.4em\relax IEEE, 2021.

\bibitem{cho2018multilingual}
J.~Cho, M.~K. Baskar, R.~Li, M.~Wiesner, S.~H. Mallidi, N.~Yalta, M.~Karafiat,
  S.~Watanabe, and T.~Hori, ``Multilingual sequence-to-sequence speech
  recognition: architecture, transfer learning, and language modeling,'' in
  \emph{2018 IEEE Spoken Language Technology Workshop (SLT)}.\hskip 1em plus
  0.5em minus 0.4em\relax IEEE, 2018, pp. 521--527.

\bibitem{yi2018adversarial}
J.~Yi, J.~Tao, Z.~Wen, and Y.~Bai, ``Adversarial multilingual training for
  low-resource speech recognition,'' in \emph{2018 IEEE International
  Conference on Acoustics, Speech and Signal Processing (ICASSP)}.\hskip 1em
  plus 0.5em minus 0.4em\relax IEEE, 2018, pp. 4899--4903.

\bibitem{ganin2016domain}
Y.~Ganin, E.~Ustinova, H.~Ajakan, P.~Germain, H.~Larochelle, F.~Laviolette,
  M.~Marchand, and V.~Lempitsky, ``Domain-adversarial training of neural
  networks,'' \emph{The journal of machine learning research}, vol.~17, no.~1,
  pp. 2096--2030, 2016.

\bibitem{nichol2018first}
A.~Nichol, J.~Achiam, and J.~Schulman, ``On first-order meta-learning
  algorithms,'' \emph{arXiv preprint arXiv:1803.02999}, 2018.

\bibitem{xiao2020adversarial}
Y.~Xiao, K.~Gong, P.~Zhou, G.~Zheng, X.~Liang, and L.~Lin, ``Adversarial meta
  sampling for multilingual low-resource speech recognition,'' in
  \emph{Association for the Advancement of Artificial Intelligence}, 2021.

\bibitem{devlin2019bert}
J.~Devlin, M.-W. Chang, K.~Lee, and K.~Toutanova, ``Bert: Pre-training of deep
  bidirectional transformers for language understanding,'' in \emph{Proceedings
  of the 2019 Conference of the North American Chapter of the Association for
  Computational Linguistics: Human Language Technologies, Volume 1 (Long and
  Short Papers)}, 2019, pp. 4171--4186.

\bibitem{baevski2019vq}
A.~Baevski, S.~Schneider, and M.~Auli, ``vq-wav2vec: Self-supervised learning
  of discrete speech representations,'' in \emph{International Conference on
  Learning Representations}, 2019.

\bibitem{baevski2020wav2vec}
A.~Baevski, Y.~Zhou, A.~Mohamed, and M.~Auli, ``wav2vec 2.0: A framework for
  self-supervised learning of speech representations,'' \emph{Advances in
  Neural Information Processing Systems}, vol.~33, 2020.

\bibitem{houlsby2019parameter}
N.~Houlsby, A.~Giurgiu, S.~Jastrzebski, B.~Morrone, Q.~De~Laroussilhe,
  A.~Gesmundo, M.~Attariyan, and S.~Gelly, ``Parameter-efficient transfer
  learning for nlp,'' in \emph{International Conference on Machine
  Learning}.\hskip 1em plus 0.5em minus 0.4em\relax PMLR, 2019, pp. 2790--2799.

\bibitem{bapna2019simple}
A.~Bapna and O.~Firat, ``Simple, scalable adaptation for neural machine
  translation,'' in \emph{Proceedings of the 2019 Conference on Empirical
  Methods in Natural Language Processing and the 9th International Joint
  Conference on Natural Language Processing (EMNLP-IJCNLP)}, 2019, pp.
  1538--1548.

\bibitem{vaswani2017attention}
A.~Vaswani, N.~Shazeer, N.~Parmar, J.~Uszkoreit, L.~Jones, A.~N. Gomez,
  {\L}.~Kaiser, and I.~Polosukhin, ``Attention is all you need,''
  \emph{Advances in Neural Information Processing Systems}, vol.~30, pp.
  5998--6008, 2017.

\bibitem{wang2018glue}
A.~Wang, A.~Singh, J.~Michael, F.~Hill, O.~Levy, and S.~R. Bowman, ``Glue: A
  multi-task benchmark and analysis platform for natural language
  understanding,'' \emph{EMNLP 2018}, p. 353, 2018.

\bibitem{sharaf2020meta}
A.~Sharaf, H.~Hassan, and H.~Daum{\'e}~III, ``Meta-learning for few-shot nmt
  adaptation,'' in \emph{Proceedings of the Fourth Workshop on Neural
  Generation and Translation}, 2020, pp. 43--53.

\bibitem{artetxe2020cross}
M.~Artetxe, S.~Ruder, and D.~Yogatama, ``On the cross-lingual transferability
  of monolingual representations,'' in \emph{Proceedings of the 58th Annual
  Meeting of the Association for Computational Linguistics}, 2020, pp.
  4623--4637.

\bibitem{li2020unsupervised}
\BIBentryALTinterwordspacing
J.~Li, R.~He, H.~Ye, H.~T. Ng, L.~Bing, and R.~Yan, ``Unsupervised domain
  adaptation of a pretrained cross-lingual language model,'' in
  \emph{Proceedings of the Twenty-Ninth International Joint Conference on
  Artificial Intelligence, {IJCAI-20}}, C.~Bessiere, Ed.\hskip 1em plus 0.5em
  minus 0.4em\relax International Joint Conferences on Artificial Intelligence
  Organization, 7 2020, pp. 3672--3678, main track. [Online]. Available:
  \url{https://doi.org/10.24963/ijcai.2020/508}
\BIBentrySTDinterwordspacing

\bibitem{lample2019cross}
\BIBentryALTinterwordspacing
A.~CONNEAU and G.~Lample, ``Cross-lingual language model pretraining,'' in
  \emph{Advances in Neural Information Processing Systems}, H.~Wallach,
  H.~Larochelle, A.~Beygelzimer, F.~d\textquotesingle Alch\'{e}-Buc, E.~Fox,
  and R.~Garnett, Eds., vol.~32.\hskip 1em plus 0.5em minus 0.4em\relax Curran
  Associates, Inc., 2019. [Online]. Available:
  \url{https://proceedings.neurips.cc/paper/2019/file/c04c19c2c2474dbf5f7ac4372c5b9af1-Paper.pdf}
\BIBentrySTDinterwordspacing

\bibitem{ye2021zero}
\BIBentryALTinterwordspacing
Q.~Ye and X.~Ren, ``Learning to generate task-specific adapters from task
  description,'' in \emph{Proceedings of the 59th Annual Meeting of the
  Association for Computational Linguistics and the 11th International Joint
  Conference on Natural Language Processing (Volume 2: Short Papers)}.\hskip
  1em plus 0.5em minus 0.4em\relax Online: Association for Computational
  Linguistics, Aug. 2021, pp. 646--653. [Online]. Available:
  \url{https://aclanthology.org/2021.acl-short.82}
\BIBentrySTDinterwordspacing

\bibitem{weller2020learning}
O.~Weller, N.~Lourie, M.~Gardner, and M.~Peters, ``Learning from task
  descriptions,'' in \emph{Proceedings of the 2020 Conference on Empirical
  Methods in Natural Language Processing (EMNLP)}, 2020, pp. 1361--1375.

\bibitem{kim2021scalable}
Y.~J. Kim, A.~A. Awan, A.~Muzio, A.~F.~C. Salinas, L.~Lu, A.~Hendy,
  S.~Rajbhandari, Y.~He, and H.~H. Awadalla, ``Scalable and efficient moe
  training for multitask multilingual models,'' 2021.

\bibitem{dong2018speech}
L.~Dong, S.~Xu, and B.~Xu, ``Speech-transformer: a no-recurrence
  sequence-to-sequence model for speech recognition,'' in \emph{2018 IEEE
  International Conference on Acoustics, Speech and Signal Processing
  (ICASSP)}.\hskip 1em plus 0.5em minus 0.4em\relax IEEE, 2018, pp. 5884--5888.

\bibitem{graves2006connectionist}
A.~Graves, S.~Fern{\'a}ndez, F.~Gomez, and J.~Schmidhuber, ``Connectionist
  temporal classification: labelling unsegmented sequence data with recurrent
  neural networks,'' in \emph{Proceedings of the 23rd international conference
  on Machine learning}, 2006, pp. 369--376.

\bibitem{vanschoren2018meta}
J.~Vanschoren, ``Meta-learning: A survey,'' \emph{arXiv preprint
  arXiv:1810.03548}, 2018.

\bibitem{vapnik1998statistical}
V.~Vapnik, ``Statistical learning theory,'' \emph{NY: Wiley}, vol.~1, p.~2,
  1998.

\bibitem{ardila2020common}
R.~Ardila, M.~Branson, K.~Davis, M.~Kohler, J.~Meyer, M.~Henretty, R.~Morais,
  L.~Saunders, F.~Tyers, and G.~Weber, ``Common voice: A massively-multilingual
  speech corpus,'' in \emph{Proceedings of The 12th Language Resources and
  Evaluation Conference}, 2020, pp. 4218--4222.

\bibitem{povey2016purely}
D.~Povey, V.~Peddinti, D.~Galvez, P.~Ghahremani, V.~Manohar, X.~Na, Y.~Wang,
  and S.~Khudanpur, ``Purely sequence-trained neural networks for asr based on
  lattice-free mmi.'' in \emph{Interspeech}, 2016, pp. 2751--2755.

\bibitem{povey2011kaldi}
\BIBentryALTinterwordspacing
D.~Povey, A.~Ghoshal, G.~Boulianne, L.~Burget, O.~Glembek, K.~N. Goel,
  M.~Hannemann, P.~Motl\'{i}\v{c}ek, Y.~Qian, P.~Schwarz, J.~Silovsk\'{y},
  G.~Stemmer, and K.~Vesel\'{y}, ``\BIBforeignlanguage{english}{The kaldi
  speech recognition toolkit},'' in
  \emph{\BIBforeignlanguage{english}{Proceedings of ASRU 2011}}.\hskip 1em plus
  0.5em minus 0.4em\relax IEEE Signal Processing Society, 2011, pp. 1--4.
  [Online]. Available: \url{https://www.fit.vut.cz/research/publication/11196}
\BIBentrySTDinterwordspacing

\bibitem{peddinti2015time}
V.~Peddinti, D.~Povey, and S.~Khudanpur, ``A time delay neural network
  architecture for efficient modeling of long temporal contexts,'' in
  \emph{Sixteenth Annual Conference of the International Speech Communication
  Association}, 2015.

\bibitem{dehak2010front}
N.~Dehak, P.~J. Kenny, R.~Dehak, P.~Dumouchel, and P.~Ouellet, ``Front-end
  factor analysis for speaker verification,'' \emph{IEEE Transactions on Audio,
  Speech, and Language Processing}, vol.~19, no.~4, pp. 788--798, 2010.

\bibitem{watanabe2018espnet}
S.~Watanabe, T.~Hori, S.~Karita, T.~Hayashi, J.~Nishitoba, Y.~Unno, N.-E.~Y.
  Soplin, J.~Heymann, M.~Wiesner, N.~Chen \emph{et~al.}, ``Espnet: End-to-end
  speech processing toolkit,'' \emph{Proc. Interspeech 2018}, pp. 2207--2211,
  2018.

\bibitem{kudo2018sentencepiece}
T.~Kudo and J.~Richardson, ``Sentencepiece: A simple and language independent
  subword tokenizer and detokenizer for neural text processing,'' \emph{EMNLP
  2018}, p.~66, 2018.

\bibitem{kingma2014adam}
D.~P. Kingma and J.~Ba, ``Adam: A method for stochastic optimization,''
  \emph{arXiv preprint arXiv:1412.6980}, 2014.

\end{thebibliography}

% \input{sec_bio}

\end{document}